\documentclass[twoside,11pt]{article}

\usepackage{blindtext}
\usepackage{float}
\usepackage[preprint]{jmlr2e}

\usepackage{tikz}
\usepackage{amsmath}
\usepackage{booktabs}
\usepackage[utf8]{inputenc}
\usepackage[T1]{fontenc}
\usepackage{float}
\usepackage{cleveref}
\usepackage{multirow}

\usepackage{lastpage}
\jmlrheading{}{}{1-\pageref{LastPage}}{}{}{}{Prashant C. Raju}

\ShortHeadings{Geometric Stability: The Missing Axis of Representations}{Raju}
\firstpageno{1}

\begin{document}

\title{Geometric Stability: The Missing Axis of Representations}

\author{\name Prashant C. Raju \email rajuprashant@gmail.com \\
       }

\editor{My editor}

\maketitle

\begin{abstract}%
Representational similarity methods compare the geometries of neural representations, but they do not measure how consistently the geometry of a single representation is recovered from subsets of its feature coordinates. We call this property \emph{geometric stability} and introduce \textit{Shesha}, which estimates it by correlating representational dissimilarity matrices from complementary random feature subsets. Shesha is not invariant to orthogonal rotations: representations with identical Gram matrices, and therefore identical linear CKA, can have different geometric stability. Controlled transformations further separate the quantities. Across $2{,}463$ encoder configurations spanning seven domains, similarity and stability are positively associated across non-PCA transformations ($\rho=+0.75$) but negatively associated under PCA-coordinate compression ($\rho=-0.47$). We further evaluate 170 pretrained vision models across six datasets. DINOv2 combines strong transfer performance with bottom-quartile stability on five of six datasets, showing that transferability and feature-split stability need not coincide. Across random feature subsets, the marginal relationship between Shesha and linear-probe variability is dataset-dependent; after controlling for task alignment with LogME, higher Shesha is associated with lower variability on five of six datasets. These results identify geometric stability as a basis-dependent property that complements representational similarity and task alignment.
\end{abstract}


%%%%%%%%%%%%%%%%%%%%%%%%%%%%%%%%%%%%%%%%%%%%%%%%%%%%%%%%%%%%%%%%%%%%%%%%%%%
\section{Introduction}
%%%%%%%%%%%%%%%%%%%%%%%%%%%%%%%%%%%%%%%%%%%%%%%%%%%%%%%%%%%%%%%%%%%%%%%%%%%
\label{sec:intro}
Characterizing internal representations is central to understanding how artificial and biological neural systems process information~\citep{bengio2013representation}. Representations from different neural systems may differ in dimensionality, feature identity, or basis, making direct comparison difficult. Representational geometry provides a way to make such comparisons through the structure they induce over samples rather than through direct correspondence between individual features. Methods such as Representational Similarity Analysis (RSA;~\citealp{Kriegeskorte2008}), Centered Kernel Alignment (CKA;~\citealp{kornblith2019similarity}), Procrustes analysis~\citep{Schnemann1966,Rohlf1990,Masarotto2018, dryden1998statistical}, and related approaches ~\citep{raghu2017svcca,morcos2018insights,Lin2024} formalize this idea in different ways, enabling comparisons across neural networks, brain representations, and other high-dimensional feature spaces. Recent work has also extended representational comparison to intrinsic Riemannian geometry while remaining invariant to hidden-state rotations~\citep{caycogajic2026geometry}.

This paper asks a different question which is how consistently is sample geometry recovered from subsets of the feature coordinates within a single representation? Similarity between two representations and consistency across feature subsets of one representation are not the same quantity. Recent work has also clarified some limits of similarity scores themselves. CKA can change substantially without corresponding changes in model behavior~\citep{davari2023reliability}; biased CKA can give spuriously high values in low-sample, high-dimensional settings~\citep{murphy2024correcting}; and high-variance directions can dominate CKA even when lower-variance directions are task-relevant~\citep{cloos2025differentiable}. At the same time, CKA and CCA have a useful functional interpretation by quantifying average alignment of optimal linear readouts across a distribution of decoding tasks~\citep{Harvey2024}. These results help specify what similarity measures capture. They do not, however, answer how the geometry of a single representation is distributed across its coordinate basis. 

That distinction matters whenever an analysis uses only part of a representation. Linear probes on selected units or features~\citep{Alain2016UnderstandingIL}, feature ablations, and other subset-based analyses depend on which coordinates are available. If complementary feature subsets induce different sample geometries, analyses performed on different subsets may also differ. This need not imply poor downstream performance, because subset sensitivity also depends on how well the available features align with the downstream task. We refer to the consistency of geometry across feature subsets as \emph{geometric stability}.

The distinction is easiest to see algebraically. For a representation matrix $X$, an orthogonal transformation $Q$ preserves the Gram matrix,
\[
(XQ)(XQ)^\top = XX^\top.
\]
Linear CKA and other rotation-invariant comparisons therefore cannot distinguish $X$ from $XQ$. Feature subsets behave differently. A subset of the rotated coordinates is generally not the rotation of the corresponding subset of the original coordinates, so the geometry induced by a feature split can change even when the full Gram matrix does not. Two representations can therefore have identical full-sample geometry while differing in how redundantly that geometry is distributed across coordinates.

We introduce \emph{Shesha} to measure this property. Shesha, named for the Hindu deity representing the invariant remainder of the cosmos~\citep{vogel1926indian-original,danielou1964hindu,dimmitt1978classical},  takes a representation $X\in\mathbb{R}^{n\times d}$ and partitions its $d$ feature coordinates into complementary subsets. It constructs a representational dissimilarity matrix (RDM) from each subset and measures the Spearman rank correlation between the resulting pairwise dissimilarities. The population quantity $\mathcal{S}(X)$ is the expected split-half correlation over balanced random partitions of the feature coordinates, and the empirical estimator $\widehat{\mathcal{S}}_K(X)$ averages this quantity over $K$ sampled partitions (Sec.~\ref{sec:framework}, Eq.~\ref{eq:shesha_full}). The construction uses the same RDM representation that underlies RSA~\citep{Kriegeskorte2008}, but applies it within a single representation rather than between two systems.

A defining property of Shesha is that it is not invariant to orthogonal transformations of the feature space (Sec.~\ref{sec:invariances} and Appendix~\ref{si:proofs}). This basis dependence is deliberate. Full cosine-distance RDMs are unchanged by an orthogonal rotation, whereas RDMs computed after selecting subsets of coordinates need not be. The resulting split-half agreement therefore measures a property that rotation-invariant similarity measures are not designed to capture. Shesha is not a replacement for similarity analysis. CKA and related measures compare representations, while Shesha measures feature-subset self-consistency within one.

We evaluate this distinction using controlled transformations across $2{,}463$ encoder configurations spanning seven domains in machine learning (audio, language, vision, and video) and in biology (protein sequences, molecular profiles, and neural population recordings). The relationship between similarity and stability depends strongly on the transformation regime. Across the non-PCA controlled transformations, the two measures are positively associated ($\rho=+0.75$), whereas under PCA-coordinate compression they are negatively associated ($\rho=-0.47$). Controlled spectral manipulations provide a complementary illustration of their different response profiles. Because CKA measures similarity between an original and a transformed representation, whereas Shesha measures feature-split consistency within the transformed representation, their numerical values are not estimates of a common quantity. Removing the leading principal component reduces CKA to $0.27$ while the modified representation retains Shesha $=0.95$; conversely, PCA-coordinate compression can preserve high CKA while substantially changing Shesha. These results show that reference-based representational similarity and feature-split geometric stability can respond differently to the same manipulation, without implying that either statistic captures a more complete notion of geometry.

We then measure geometric stability in 170 pretrained vision models across six datasets. DINOv2 provides a useful case. Despite strong transfer performance, it lies in the bottom quartile of geometric stability on five of the six datasets. This pattern is not reducible to effective dimensionality alone. Participation-ratio analyses show that low Shesha can coexist with high effective dimensionality. Other model families occupy different parts of the stability--transferability space, and contrastively trained models often combine strong transfer with high stability.

Finally, we test whether geometric stability is related to the reliability of linear probes trained on random feature subsets. The marginal relationship between Shesha\textsubscript{FS} and probe variability is heterogeneous across datasets, showing that geometric stability alone does not determine downstream subset sensitivity. Because probe behavior also depends strongly on task alignment~\citep{njaradi2026optimal}, we condition on LogME. After controlling for LogME, higher Shesha\textsubscript{FS} is associated with lower probe variability on five of six datasets, while Flowers-102 shows no significant conditional association. Conversely, LogME remains strongly negatively associated with probe variability after controlling for Shesha\textsubscript{FS} on all six datasets. Thus, task alignment is the more consistent correlate of subset-probe reliability, while geometric stability provides complementary information about sensitivity to which feature coordinates are available.

Taken together, the results support geometric stability as a basis-dependent property of representations that is distinct from similarity, transfer performance, and task alignment. We observe this distinction across artificial models and biological representations, but do not treat stability as a universal measure of representation quality. Its intended use is narrower, which is to quantify whether the geometry of a representation is recoverable from subsets of its learned coordinates, and to identify cases in which subset-based analyses may be sensitive to the particular coordinates selected.

%%%%%%%%%%%%%%%%%%%%%%%%%%%%%%%%%%%%%%%%%%%%%%%%%%%%%%%%%%%%%%%%%%%%%%%%%%%
\section{The Geometric Stability Framework}
%%%%%%%%%%%%%%%%%%%%%%%%%%%%%%%%%%%%%%%%%%%%%%%%%%%%%%%%%%%%%%%%%%%%%%%%%%%
\label{sec:framework}
We introduce geometric stability, a property of a single representation that measures how reliably its pairwise distance geometry can be recovered from complementary halves of its feature dimensions. The subsections below define the Shesha\textsubscript{FS} population functional and its Monte Carlo estimator, and characterize the invariances it does and does not inherit.

\begin{figure*}[!h]
\includegraphics[width=\textwidth]{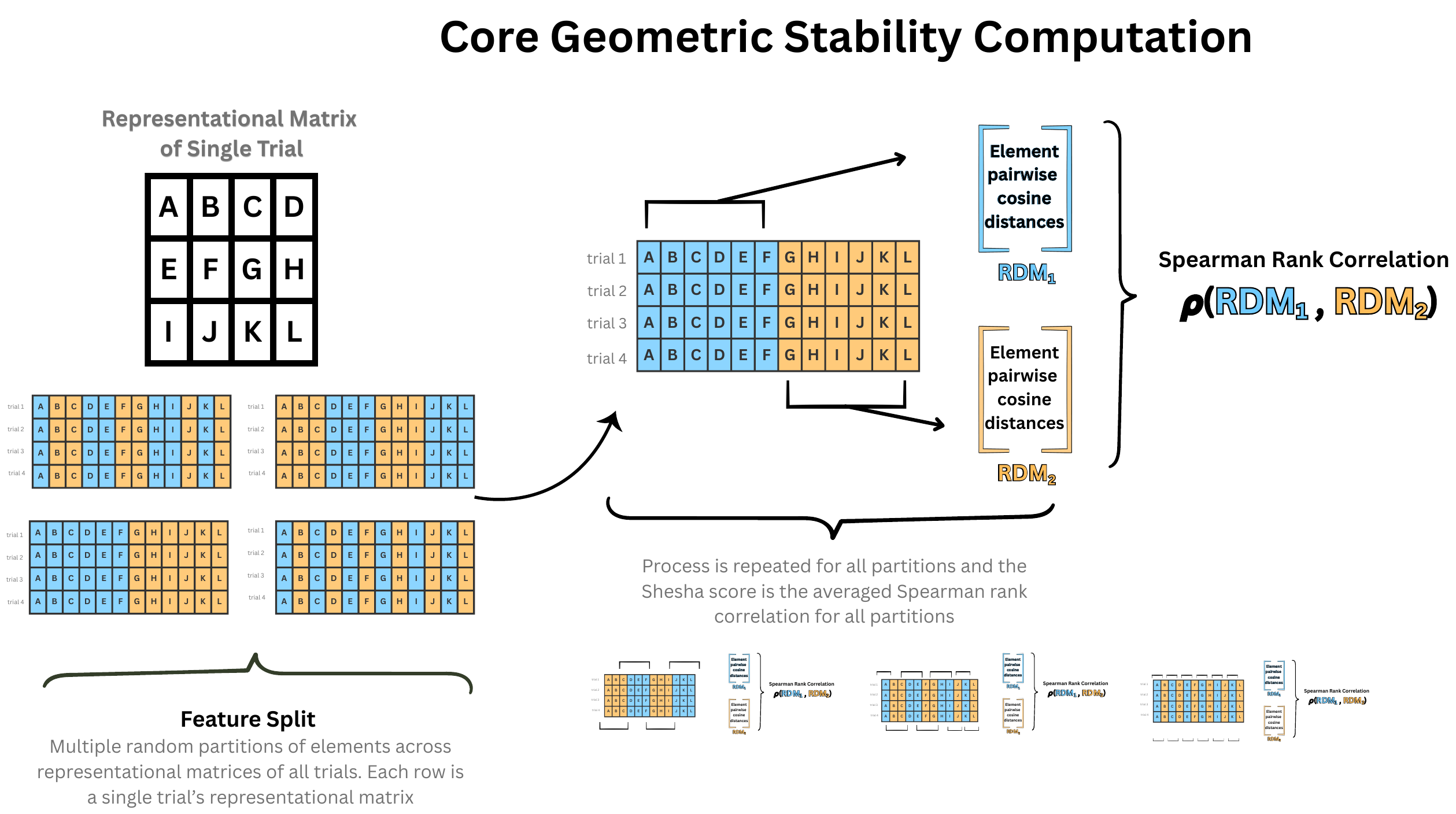}
\caption{Illustration of Shesha\textsubscript{FS} Computation.}
\label{fig:conceptual-illustration}
\end{figure*}

\subsection{Formal Definition}

Let $X \in \mathbb{R}^{n \times d}$ be a representation matrix of $n$ samples in $d$ dimensions. Let $\Pi_d$ denote the uniform distribution over balanced complementary partitions $(A,B)$ of the feature index set $\{1,\ldots,d\}$, with $|A|=\lfloor d/2\rfloor$ and $|B|=\lceil d/2\rceil$. For each half, we construct an RDM using pairwise cosine distances:
\begin{equation}
    D^{(s)}_{ij} = 1 - \frac{x_i^{(s)} \cdot x_j^{(s)}}
    {\|x_i^{(s)}\|\,\|x_j^{(s)}\|},
    \quad s \in \{A,B\},
    \label{eq:rdm}
\end{equation}
where $x_i^{(s)}$ and $x_j^{(s)}$ denote the subvectors of samples $i$ and $j$ restricted to the dimensions in half $s$. We define feature-split geometric stability as the expected Spearman rank correlation between the vectorized upper triangles of the two complementary half-RDMs:
\begin{equation}
    \mathcal{S}(X)
    = \mathbb{E}_{(A,B)\sim\Pi_d}
    \left[
    \rho_s\!\left(
        \operatorname{vec}_{\triangle}(D^{(A)}),\;
        \operatorname{vec}_{\triangle}(D^{(B)})
    \right)
    \right].
    \label{eq:shesha_full}
\end{equation}
In practice, we estimate $\mathcal{S}(X)$ by Monte Carlo sampling of $K$ independent balanced partitions:
\begin{equation}
    \widehat{\mathcal{S}}_{K}(X)
    = \frac{1}{K} \sum_{k=1}^{K}
    \rho_s\!\left(
        \operatorname{vec}_{\triangle}(D^{(A_k)}),\;
        \operatorname{vec}_{\triangle}(D^{(B_k)})
    \right),
    \qquad (A_k,B_k) \overset{\mathrm{iid}}{\sim} \Pi_d .
    \label{eq:shesha_estimator}
\end{equation}
$\mathcal{S}(X) \in [-1,1]$, with $\mathcal{S}(X) \approx 1$ indicating that complementary feature subsets recover the same pairwise geometry, $\mathcal{S}(X) \approx 0$ indicating little consistent rank agreement between the two half-RDMs, and negative values indicating systematically opposing pairwise-distance rankings. Unless otherwise specified, all reported Shesha\textsubscript{FS} values use the Monte Carlo estimator in Eq.~\eqref{eq:shesha_estimator} with $K=30$. The use of Spearman correlation makes $\mathcal{S}$ invariant to strictly monotone rescaling of the RDM distance values, and the use of cosine distance makes it invariant to uniform nonzero scaling of the representation. The full invariance structure is established formally in Sec.~\ref{sec:invariances}.

\subsection{Basis Interpretation}
\label{sec:spectral}

The feature-split procedure in Eq.~\eqref{eq:shesha_full} asks whether random halves of the coordinate axes recover the same pairwise geometry. What it measures is how geometric information is distributed across the coordinate basis of $X$, not a property of the eigenspectrum alone.

To see the distinction, write the covariance $\Sigma = \frac{1}{n} X^\top X$ as $\Sigma = V \Lambda V^\top$. Two second-order properties can influence split-half agreement: the eigenspectrum $\Lambda$, which describes how variance is distributed across directions, and the orientation $V$, which describes how those directions map onto the $d$ coordinates that a partition splits. However, $\mathcal{S}$ is not determined by $\Sigma$: it also depends on the sample configuration beyond second-order statistics.

When variance is concentrated in a few coordinates, a random partition tends to yield halves with asymmetric information content, and the resulting half-RDMs can disagree, producing lower $\mathcal{S}$. When relevant structure is distributed redundantly across coordinates so that each half recovers similar pairwise structure, $\mathcal{S}$ tends to be higher. These are tendencies rather than consequences of the eigenspectrum alone: an orthogonal rotation leaves $\Lambda$ unchanged while redistributing structure across coordinates, and can change $\mathcal{S}$ (Sec.~\ref{sec:invariances}).

So \(\mathcal{S}\) is not determined by the eigenspectrum and is distinct from basis-invariant spectral summaries such as spectral entropy and participation ratio. A representation can have a broad eigenspectrum yet a low $\mathcal{S}$ when its variance is spread non-redundantly, so that no half recovers the whole. The spectral-deletion experiments in Sec.~\ref{sec:results} show that $\mathcal{S}$ and CKA can respond differently as components are removed. We interpret this dissociation as evidence that the two statistics measure different properties of a representation. CKA compares the modified representation with a reference representation and $\mathcal{S}$ measures internal agreement between complementary feature subsets of the modified representation. The dissociation alone does not imply that either statistic captures a more complete notion of geometry.

\subsection{Invariance Properties}
\label{sec:invariances}

The invariance structure of $\mathcal{S}$ overlaps with that of linear CKA in some respects and departs from it in one property central to this work: $\mathcal{S}$ is not invariant to orthogonal changes of the feature basis. This departure provides the formal basis for the empirical dissociations reported in Sec.~\ref{sec:results}. Full proofs are given in Appendix~\ref{si:proofs}.

\subsubsection{Uniform Scalar Scaling}
Cosine distance is unchanged by multiplying all feature coordinates by the same nonzero scalar, so $\mathcal{S}(\alpha X) = \mathcal{S}(X)$ for any $\alpha \neq 0$, whenever the cosine distances are defined. Linear CKA shares this invariance.

\subsubsection{Feature Permutation}
Relabeling coordinate indices does not change representational content. Since partitions are drawn uniformly at random over coordinate indices, the distribution over partitions is exchangeable under permutation, and $\mathcal{S}$ is invariant. CKA is invariant for the same reason as orthogonal rotation: permutation matrices are orthogonal, so $XX^\top$ is preserved.

\subsubsection{PCA Coordinate Compression}
Let $V_r \in \mathbb{R}^{d\times r}$ contain the top $r$ principal directions of $X$, and let
\begin{equation}
    Z_r = X V_r
\end{equation}
denote the representation expressed in the retained PCA coordinates. This operation is distinct from the rank-$r$ reconstruction $X V_r V_r^\top$, which is expressed back in the original $d$-dimensional coordinate basis. Because Shesha\textsubscript{FS} is explicitly basis-dependent, these two representations need not have the same stability.

Shesha\textsubscript{FS} is not invariant to PCA coordinate compression: there exist representations for which
\begin{equation}
    \mathcal{S}(X V_r) \neq \mathcal{S}(X).
\end{equation}
Appendix~\ref{si:proofs-pca} gives an explicit counterexample in which $r=\operatorname{rank}(X)$, so PCA coordinate compression discards no nonzero singular directions and preserves the sample Gram matrix exactly,
\begin{equation}
    (X V_r)(X V_r)^\top = X X^\top,
\end{equation}
while changing $\mathcal{S}$. Thus the result is a statement of non-invariance, not a claim that PCA compression must decrease Shesha for every representation.

In the empirical PCA interventions in Sec.~\ref{sec:results}, Shesha\textsubscript{FS} decreases as progressively more principal-component coordinates are retained, while CKA to the uncompressed representation rises toward its full-rank value. Because retaining more coordinates discards fewer dimensions, this trend is not what an account based on information loss alone would predict. More generally, lower Shesha in PCA coordinates should not be attributed to discarding dimensions: the lossless counterexample in Appendix~\ref{si:proofs-pca} shows that re-expression in PCA coordinates can change $\mathcal{S}$ even when all nonzero singular directions are retained.

\subsubsection{Orthogonal Rotation}
Let $Y = XQ$ for $Q \in \mathcal{O}(d)$. Then $YY^\top = XQQ^\top X^\top = XX^\top$, so the Gram matrix is preserved and linear $\operatorname{CKA}(X, Y) = 1$ for any orthogonal $Q$. Shesha\textsubscript{FS}, by contrast, is not invariant: $Q$ redistributes geometric information across the coordinate axes, so a random feature partition of $Y$ captures different directions than the same partition of $X$, and $\mathcal{S}(Y)$ can differ from $\mathcal{S}(X)$. This is the key formal dissociation: linear CKA is invariant to orthogonal changes of feature basis, whereas Shesha\textsubscript{FS} need not be, since an orthogonal $Q$ can change $\mathcal{S}$ while holding CKA fixed. As a direct demonstration, rotating a stable representation into its own eigenbasis, an orthogonal transformation that preserves the eigenspectrum and rank exactly, collapses Shesha\textsubscript{FS} from $0.903$ to near zero while leaving CKA at $1.000$. The constructive counterexample and this numerical demonstration are given in Appendix~\ref{si:proof-ortho-transform}.

\subsubsection{Strictly Monotone RDM Transformations}
Spearman rank correlation depends only on ranks. Applying the same strictly increasing transformation to both half-RDMs preserves their ranks, while applying the same strictly decreasing transformation reverses both rank orderings; in either case their Spearman correlation is unchanged. Thus $\mathcal{S}$ is invariant to a common strictly monotone transformation applied directly to the RDM entries. This is an RDM-level invariance and is not directly applicable to feature-space metrics such as linear CKA.

\subsection{Connection to the RSA Noise Ceiling}
\label{sec:noise_ceiling}

Shesha is conceptually related to the RSA noise ceiling~\citep{Nili2014}, but the two quantify different forms of reliability. In a standard RSA noise-ceiling analysis, each subject or replicate contributes an RDM computed from its \textit{full} available feature population, and agreement across these RDMs reflects the reproducibility of the estimated representational geometry under measurement variability.

Shesha instead \textit{partitions} the feature population itself. Each half-RDM is computed from only a complementary subset of the coordinates of a single representation, so $\mathcal{S}$ measures how consistently the pairwise geometry is recovered when different parts of the feature basis are retained. Thus the noise ceiling asks whether the full representational geometry is reproducible across measurements and Shesha asks whether that geometry is recoverable from subsets of the feature population. The connection is therefore methodological rather than an identity of procedures (Appendix~\ref{si:noise_ceiling}).

%%%%%%%%%%%%%%%%%%%%%%%%%%%%%%%%%%%%%%%%%%%%%%%%%%%%%%%%%%%%%%%%%%%%%%%%%%%
\section{Distinctness of Stability and Similarity}
%%%%%%%%%%%%%%%%%%%%%%%%%%%%%%%%%%%%%%%%%%%%%%%%%%%%%%%%%%%%%%%%%%%%%%%%%%%
\label{sec:results}
Section~\ref{sec:invariances} established formally that Shesha\textsubscript{FS} can change under transformations that leave the full Gram matrix, and therefore linear CKA, unchanged. We now examine how this distinction appears empirically. The analyses below compare the responses of Shesha\textsubscript{FS} and several representational similarity measures under controlled transformations, across pretrained encoders, and under a controlled training intervention using SAM. Together, these experiments test whether feature-split stability captures variation that is not reducible to standard representational similarity.

\subsection{Construct Validation}
\label{sec:validation}

We first test whether $\mathcal{S}$ responds systematically to controlled changes in feature-split structure and whether stability and similarity can vary separately. In synthetic representations generated by mixing a shared low-rank signal with isotropic noise, Shesha is nearly monotonic with the signal-to-noise mixing parameter $\alpha \in [0,1]$ ($\rho = 0.997$, $p < 10^{-86}$; Appendix~\ref{si:ground-truth}). A separate balanced construction samples all four quadrants of the stability--similarity plane, including cases with CKA $>0.97$ and low $\mathcal{S}$ (Appendix~\ref{app:balanced-quad}). These examples show that high similarity does not imply high feature-split stability, and that high stability does not require high similarity.

\subsection{Spectral Manipulations Separate Similarity from Feature-Split Stability}
\label{sec:mechanism}

\begin{figure*}[ht]
  \includegraphics[width=\textwidth]{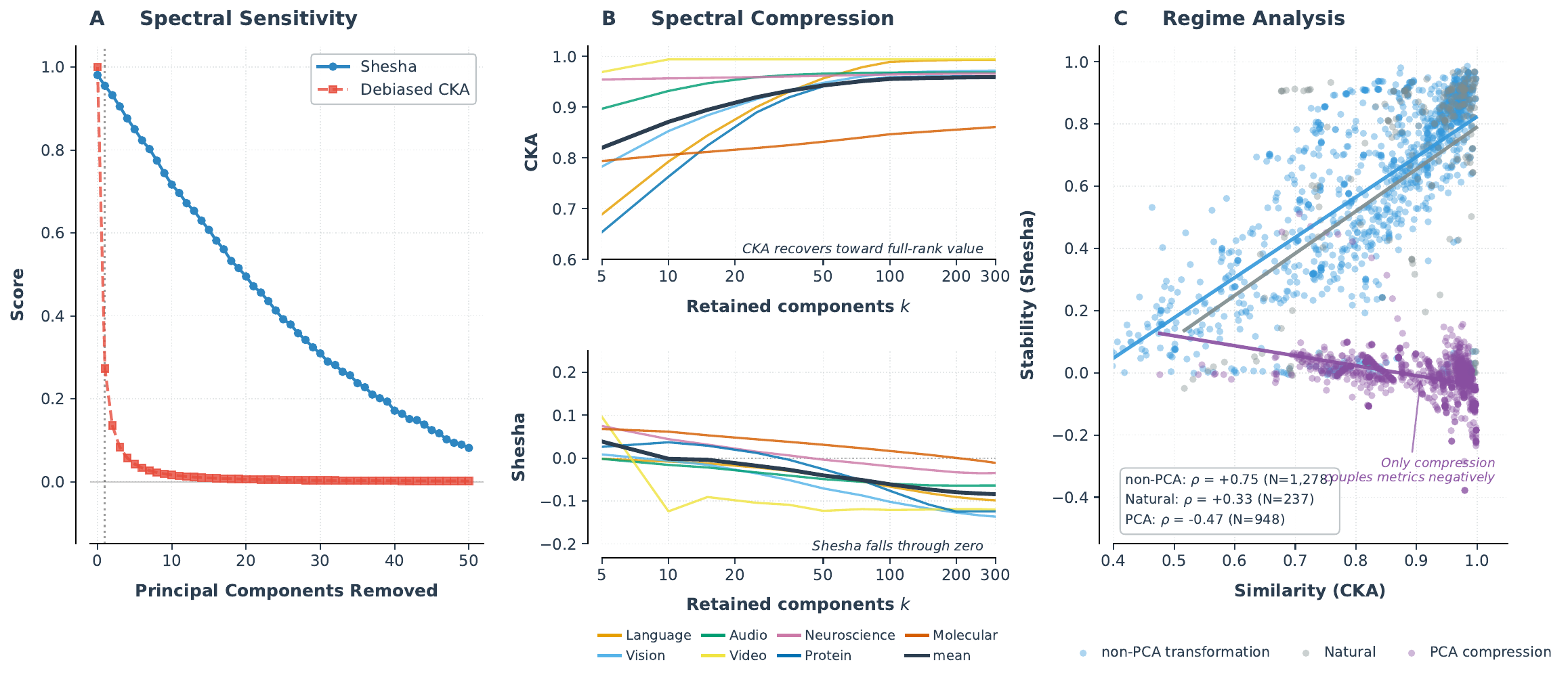}
\caption{Spectral manipulations separate similarity from feature-split stability. Panels A and B apply complementary spectral transformations. A.~Spectral deletion: after removal of the leading principal component, CKA between the modified and original representation falls to $0.27$, while Shesha\textsubscript{FS} of the modified representation remains at $0.95$ and declines more gradually as additional components are removed. B.~PCA coordinate compression: as the retained dimension $k$ increases, CKA to the original representation rises toward its full-rank value, whereas Shesha\textsubscript{FS} decreases across all seven domains (curves are per-domain means). These quantities answer different questions: CKA measures similarity between the original and transformed representations, whereas Shesha\textsubscript{FS} measures agreement between complementary feature subsets within the transformed representation. C.~Regime analysis across $2{,}463$ configurations in seven domains. The metrics are positively associated for the non-PCA transformation regime used here ($\rho = +0.75$, $N = 1{,}278$), weakly associated across natural encoders ($\rho = +0.33$, $N = 237$), and negatively associated under PCA coordinate compression ($\rho = -0.47$, $N = 948$). A pooled correlation is not interpreted because it depends on the mixture of regimes. Panel A is replicated with debiased CKA, Procrustes, and PWCKA in Appendix Fig.~\ref{fig:spectral}.}
\label{fig:evidence}
\end{figure*}

The formal result in Sec.~\ref{sec:invariances} shows that Shesha\textsubscript{FS} is not determined by the full Gram matrix: an orthogonal change of basis can leave $XX^\top$ unchanged while changing $\mathcal{S}$. The experiments here ask how that distinction appears under controlled spectral transformations. Importantly, the two quantities are not being treated as alternative estimates of the same target. CKA compares two representations, whereas Shesha\textsubscript{FS} measures the internal agreement between complementary feature subsets of one representation. The purpose of the following analyses is therefore to characterize where their responses agree and where they separate.

We first remove leading principal components from synthetic representations $X = U \Sigma V^\top$, in which $U$ and $V$ are random orthonormal matrices, so that the principal directions are mixed across the coordinate axes, and the singular values decay as $\sigma_i = 100/i$ ($i = 1, 2, \ldots$), giving covariance eigenvalues proportional to $i^{-2}$ (Appendix~\ref{si:spectral-deletion}). CKA is computed between the original and modified representations, while Shesha\textsubscript{FS} is computed within the modified representation. Removing the leading component reduces linear CKA from $1.0$ to $0.27$, whereas Shesha\textsubscript{FS} is essentially unchanged ($0.981$ to $0.955$) and declines gradually as additional components are removed (Fig.~\ref{fig:evidence}A). Procrustes and PWCKA respond like CKA (Appendix Fig.~\ref{fig:spectral}). The CKA response is consistent with its variance weighting: changes along high-variance directions dominate the Gram matrix, so deleting the leading component registers as a large departure from the reference~\citep{cloos2025differentiable}. The small initial change in Shesha\textsubscript{FS} is consistent with the random orientation of the principal directions across coordinates. Because each coordinate contains contributions from multiple principal directions, removing one direction can leave the two complementary feature subsets with similar pairwise geometries. The two statistics answer different questions (Sec.~\ref{sec:spectral}): CKA quantifies how much the modified representation differs from the original reference, whereas Shesha\textsubscript{FS} quantifies feature-split consistency within the modified representation. The pattern holds across raw, centered, normalized, and whitened preprocessing: at $k=30$, debiased CKA is near zero or negative in every condition while Shesha\textsubscript{FS} remains positive ($0.28$ to $0.42$), and whitening lowers the Shesha\textsubscript{FS} baseline ($0.98$ to $0.50$ at $k=0$), consistent with noise amplification when variance is equalized across directions (Appendix~\ref{si:preprocessing}). A complementary control injects Gaussian noise only into components outside the subspace explaining $90\%$ of the variance: small perturbations leave CKA above $0.999$ and Shesha\textsubscript{FS} near its baseline of $0.971$, while larger perturbations change both, so the dissociation seen under leading-component deletion is not reproduced by tail perturbation (Appendix~\ref{si:tail_noise}).

PCA coordinate compression provides a complementary manipulation. For each retained dimension $k$, we represent the data in the coordinates of the top $k$ principal components and compare that compressed representation with the original using CKA. As $k$ increases, CKA rises toward its full-rank value, while Shesha\textsubscript{FS} decreases (Fig.~\ref{fig:evidence}B). In vision representations, retaining $k=300$ components yields CKA $=0.972$ while Shesha\textsubscript{FS} changes from $+0.732$ in the uncompressed representation to $-0.136$ in PCA coordinates; in language, the corresponding values are CKA $=0.993$ and Shesha\textsubscript{FS} $=-0.098$ from an uncompressed value of $+0.761$. Across the PCA sweep, retained dimension correlates positively with CKA (vision $+0.84$, language $+0.91$, audio $+0.79$) and negatively with Shesha\textsubscript{FS} (vision $-0.86$, language $-0.94$, audio $-0.88$; negative in all seven domains). Because PCA orders coordinates by variance and decorrelates them, complementary coordinate subsets can contain different mixtures of variance scales. The resulting split-half geometries need not agree, even as the compressed representation becomes increasingly similar to the original as more components are retained. The formal result in Appendix~\ref{si:proofs} establishes only non-invariance to PCA coordinate compression; the monotonic decreases observed here are empirical properties of these representations, not a general theorem.

The same distinction appears across the broader transformation suite (Fig.~\ref{fig:evidence}C). In the non-PCA transformation regime used here, which includes random projection together with the corresponding feature-subset, feature-selection, and noise controls, the two metrics are positively associated ($\rho = +0.75$, $N = 1{,}278$). Random projection has a direct distance-preservation guarantee under the Johnson--Lindenstrauss lemma~\citep{Johnson1984,Dasgupta2002}; the other transformations are included empirically because they preserve the relevant pairwise structure closely in our parameter ranges (Appendix Table~\ref{si:regime-analysis}). Across natural encoders, the association is weaker ($\rho = +0.33$, $N = 237$). PCA coordinate compression is the only regime with a negative association ($\rho = -0.47$, $N = 948$). These regime-specific results are more informative than a pooled correlation because the pooled value depends on how many configurations are drawn from each transformation family.

Across all $2{,}463$ encoder configurations spanning vision, language, audio, video, neuroscience, proteins, and molecular data, the same regime dependence is visible within individual domains (Appendix Table~\ref{tab:domains}). A mixed-effects model with base-model identity as a grouping factor gives an ICC of $0.10$, indicating that base-model identity accounts for a modest fraction of the observed stability variance. We therefore do not use encoder identity, or a pooled cross-domain correlation, as evidence for distinctness. The main evidence is the formal basis dependence in Sec.~\ref{sec:invariances}, the controlled spectral manipulations above, and the regime-specific associations.

\subsection{SAM Perturbation Separates Stability and Similarity}
\label{sec:sam_ablation}
To test whether a controlled change in training can alter stability and similarity differently, we trained ResNet-18 models on CIFAR-10 and CIFAR-100 under identical conditions while varying only the Sharpness-Aware Minimization (SAM;~\citealt{foret2021sharpnessaware}) perturbation radius $\rho \in \{0,0.01,0.02,0.05,0.1,0.2\}$, with $\rho=0$ corresponding to standard SGD. Each configuration was run over 15 random seeds. Table~\ref{tab:sam_ablation} reports the mean and standard deviation of each metric across seeds.

Test accuracy varies little across the sweep (94.91--95.56\% on CIFAR-10; 76.96--78.05\% on CIFAR-100), while CKA between each SAM model and its SGD baseline decreases as $\rho$ increases, reaching 0.925 on CIFAR-10 and 0.765 on CIFAR-100. Shesha\textsubscript{FS} does not follow the same trajectory. On CIFAR-10 it rises from 0.806 under SGD to 0.872 at $\rho=0.05$ and $0.10$, then falls to 0.851 at $\rho=0.2$. On CIFAR-100 it rises more gradually, from 0.805 under SGD to 0.822 at $\rho=0.2$. At the largest observed increase for each dataset, Shesha\textsubscript{FS} exceeds the matched SGD value on all 15 seeds (CIFAR-10, $\rho=0.05$: $+0.067$, paired $t_{14}=25.3$, $p<10^{-12}$; CIFAR-100, $\rho=0.2$: $+0.018$, $t_{14}=16.6$, $p<10^{-9}$; two-sided).

This ablation provides a controlled example in which representational similarity to the SGD baseline and feature-split stability move differently while test accuracy remains in a narrow range. It does not imply that SAM generally increases stability, nor that lower CKA should correspond to higher Shesha\textsubscript{FS}. Rather, it shows that the two quantities can respond differently to the same training intervention. The supervised Shesha variants and full per-seed results are reported in Appendix~\ref{si:sam_ablation}.

\begin{figure*}[ht]
  \includegraphics[width=\textwidth]{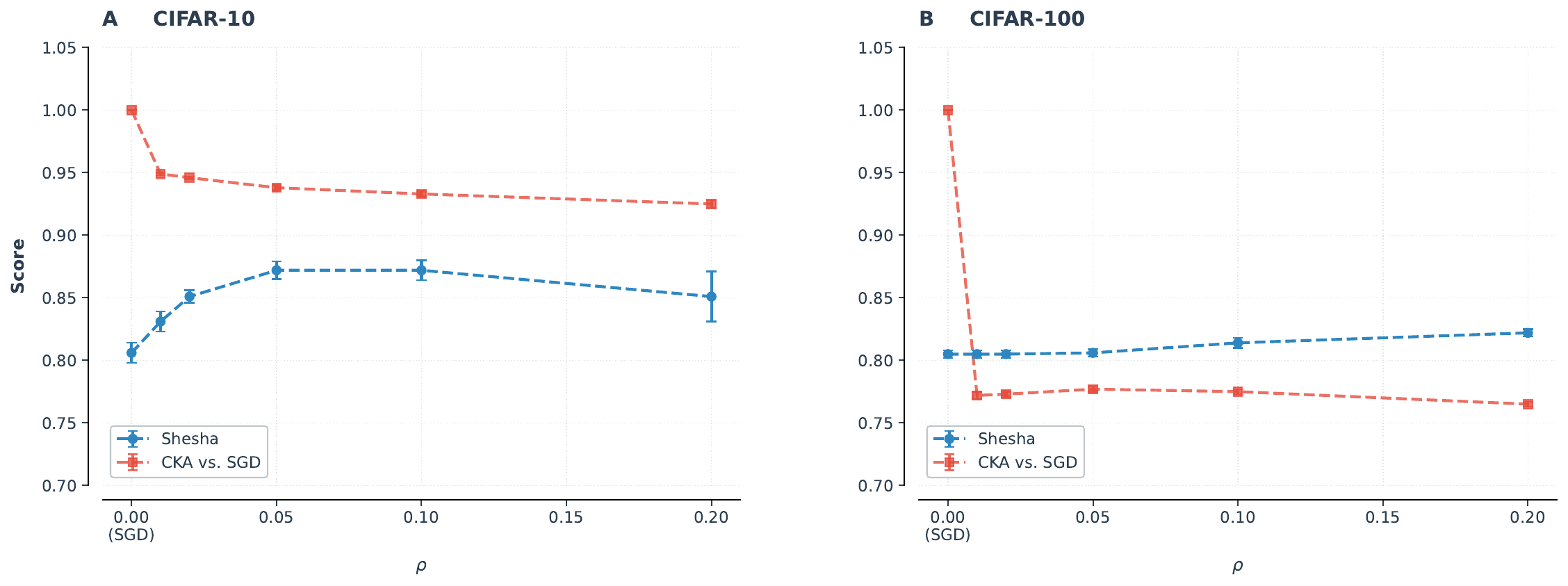}
\caption{SAM perturbation radius ablation. ResNet-18 models were trained on CIFAR-10 and CIFAR-100 with identical hyperparameters while varying only the SAM perturbation radius $\rho$ ($\rho=0$ is standard SGD). Points are means over 15 seeds; error bars are $\pm1$ SD (smaller than the markers for CKA). As $\rho$ increases, CKA to the SGD baseline decreases, while Shesha\textsubscript{FS} follows a different trajectory. Values are reported in Table~\ref{tab:sam_ablation}.}
\label{fig:sam_ablation}
\end{figure*}

\begin{table}[H]
\centering
\caption{SAM perturbation radius ablation. Values are mean\,$\pm$\,SD over 15 seeds. Test accuracy remains in a narrow range and CKA to the SGD baseline decreases across the sweep, while Shesha\textsubscript{FS} follows a different trajectory.}
\label{tab:sam_ablation}
\begin{tabular}{lcccc}
\hline
& \multicolumn{2}{c}{CIFAR-10} & \multicolumn{2}{c}{CIFAR-100} \\
$\rho$ & CKA vs.\ SGD & Shesha\textsubscript{FS} & CKA vs.\ SGD & Shesha\textsubscript{FS} \\
\hline
0.00 (SGD) & 1.000 & $0.806 \pm 0.008$ & 1.000 & $0.805 \pm 0.003$ \\
0.01       & 0.949 & $0.831 \pm 0.008$ & 0.772 & $0.805 \pm 0.003$ \\
0.02       & 0.946 & $0.851 \pm 0.005$ & 0.773 & $0.805 \pm 0.003$ \\
0.05       & 0.938 & $0.872 \pm 0.007$ & 0.777 & $0.806 \pm 0.003$ \\
0.10       & 0.933 & $0.872 \pm 0.008$ & 0.775 & $0.814 \pm 0.004$ \\
0.20       & 0.925 & $0.851 \pm 0.020$ & 0.765 & $0.822 \pm 0.003$ \\
\hline
\end{tabular}
\end{table}

%%%%%%%%%%%%%%%%%%%%%%%%%%%%%%%%%%%%%%%%%%%%%%%%%%%%%%%%%%%%%%%%%%%%%%%%%%%
\section{Geometric Stability in Pretrained Vision Models}
%%%%%%%%%%%%%%%%%%%%%%%%%%%%%%%%%%%%%%%%%%%%%%%%%%%%%%%%%%%%%%%%%%%%%%%%%%%
\label{sec:vision}
High transferability does not necessarily imply that representational structure is redundantly distributed across feature coordinates. We test the relationship across 170 pretrained vision models from 36 architectural families, evaluated on six datasets spanning four visual domains: natural images (CIFAR-10 and CIFAR-100; ~\citealt{Krizhevsky09learningmultiple}), fine-grained recognition (Flowers 102;~\citealt{Nilsback08}; Oxford Pets;~\citealt{parkhi12a}), texture (DTD;~\citealt{cimpoi14describing}), and remote sensing (EuroSAT;~\citealt{helber2018introducing, Helber2019}). Transferability is estimated with LogME~\citep{you2021logme, you_ranking_2022}, a label-efficient proxy for linear-probe performance. To check that the stability rankings are not artifacts of a single feature-partition seed, we recomputed the full CIFAR-10 sweep under three seeds (9, 320, and 1991) for all 170 models. Shesha\textsubscript{FS} is highly reproducible across seeds (Spearman $\rho \geq 0.993$, median per-model CV 0.75\%; Appendix~\ref{si:seed}).

\begin{figure}[h]
\centering
\includegraphics[width=\textwidth]{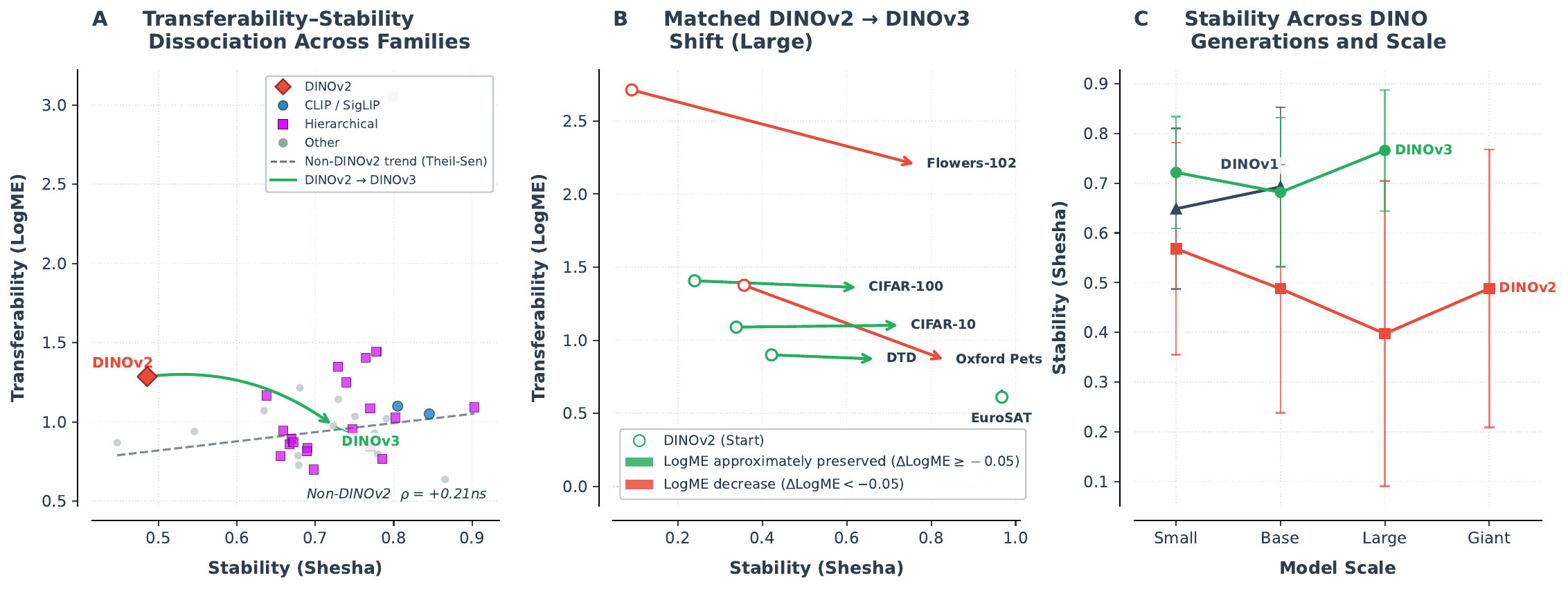}
\caption{Transferability and geometric stability across vision models.
A. Family-mean transferability (LogME) against geometric stability
(Shesha\textsubscript{FS}) for 36 architectural families, averaged across the
six datasets. DINOv2 combines relatively high transferability with unusually
low stability, illustrating a dissociation between the two quantities rather
than a general trade-off. Across the remaining 35 families, the association is
weakly positive (Spearman $\rho=+0.21$, not significant), while CLIP and
SigLIP combine relatively high stability with competitive transferability.
The arrow marks the aggregate shift from DINOv2 to DINOv3. The per-dataset
breakdown, including the EuroSAT case in which DINOv2 is itself highly stable,
is shown in Appendix~\ref{fig:family_grid}.
B. Matched large-scale DINOv2 and DINOv3 models on the
stability--transferability plane, with one arrow per dataset. DINOv3 generally
moves toward higher Shesha\textsubscript{FS}, whereas the corresponding change
in LogME is dataset-dependent. Four datasets show higher stability with LogME
approximately preserved ($\Delta\mathrm{LogME}\geq-0.05$), while Flowers-102
and Oxford Pets show larger decreases in LogME. The panel therefore
illustrates a generational shift in geometric stability without implying a
universal stability--transferability trade-off.
C. Geometric stability as a function of model scale, shown as
mean $\pm$ SD across the six datasets. DINOv2 stability decreases from small
to large models and increases again at giant scale, whereas DINOv3 is
comparatively stable across the available scales. Participation-ratio analyses
in Sec.~\ref{sec:not_eff_dim} show that these changes are not tracked by
effective dimensionality alone.}
\label{fig:dino}
\end{figure}

\subsection{DINOv2 Separates Transferability from Stability}
\label{sec:dinov2}

The family rankings make the DINOv2 pattern clear (Table~\ref{tab:dino}; Appendix Fig.~\ref{fig:family_grid}). DINOv2~\citep{oquab2024dinov} ranks first or second in transferability on three of six datasets (LogME rank 1/36 on Flowers-102 and 2/36 on CIFAR-10 and CIFAR-100), while ranking last or next-to-last in geometric stability on those same datasets (36/36, 35/36, and 36/36, respectively). It also falls in the bottom quartile on Oxford Pets (33/36) and DTD (29/36). EuroSAT is the exception: DINOv2 is both highly transferable and highly stable there (Shesha\textsubscript{FS} $=0.950$, rank 4/36). The CIFAR-10 ordering is stable across feature-partition seeds, with DINOv2 retaining the lowest family-mean Shesha\textsubscript{FS} under all three seeds (Appendix~\ref{si:seed}). On EuroSAT, DINOv2 is also much more spectrally concentrated than on CIFAR-10 (top-eigenvalue share 0.206 vs.\ 0.045; participation ratio 16.1 vs.\ 99.0), consistent with the benchmark-wide associations reported in Sec.~\ref{sec:not_eff_dim}.

DINOv2 is therefore useful as a counterexample to a simple transfer--stability tradeoff: high transferability can coexist with unusually low feature-split stability. The pattern is not explained by low effective dimensionality. On the natural-image datasets, DINOv2 has among the highest participation ratios in the benchmark while remaining among the least stable (Sec.~\ref{sec:not_eff_dim}). Because participation ratio is rotation-invariant whereas Shesha\textsubscript{FS} depends on the learned coordinate basis, the two quantities need not move together. The PCA-coordinate analysis in Appendix~\ref{si:proofs-pca} establishes this distinction formally, and the optimizer ablation in Sec.~\ref{sec:sam_ablation} provides a controlled empirical example in which stability changes without a corresponding change in the comparison metric.

\subsection{DINOv3 Shows Higher Geometric Stability}
\label{sec:dinov3}

DINOv3~\citep{simeoni2025dinov3} is useful for asking whether the DINOv2 pattern persists in the next model generation. Among other training changes, DINOv3 introduces Gram anchoring to reduce dense-feature degradation during long training runs. Across the six datasets, DINOv3 has higher mean Shesha\textsubscript{FS} than DINOv2 (0.722 vs.\ 0.485) but lower mean transferability (LogME 0.976 vs.\ 1.286; Table~\ref{tab:dino}). The difference is not uniform across datasets or scales. From small to large, DINOv2 stability falls from 0.571 to 0.402, whereas DINOv3 remains comparatively stable (0.721 to 0.763). At matched large scale, DINOv3 has higher stability with little or no transfer reduction on four of six datasets. On CIFAR-10, for example, Shesha\textsubscript{FS} rises from 0.339 to 0.730 while LogME is similar (1.089 vs.\ 1.102). Flowers-102 and Oxford Pets show a larger transfer reduction. Matching scale also narrows the mean transfer gap (LogME 1.349 for DINOv2 vs.\ 1.183 for DINOv3), indicating that part of the family-average difference reflects the unmatched DINOv2 giant variant. Register variants of DINOv2 show slightly lower stability than their non-register counterparts at each scale (for example, 0.392 vs.\ 0.402 at large).

These comparisons show that the low-stability DINOv2 pattern is not shared by DINOv3. They do not, however, isolate Gram anchoring as the cause: the generations differ in several aspects of training and model design. The DINOv3 result is therefore consistent with, but does not establish, a contribution from Gram anchoring or other training changes that affect the distribution of information across coordinates.

\subsection{Architectural and Training Associations}
\label{sec:determinants}

Geometric stability also varies across architecture and training families. CLIP-family models~\citep{radford2021learning} have higher stability than the self-supervised group on all six datasets (Mann-Whitney $p<0.05$ for each), and EVA-02, which reconstructs CLIP features rather than raw pixels, ranks among the more stable models on most benchmarks. These results associate contrastive or feature-aligned training with higher stability, but the benchmark is observational and does not isolate the training target from other differences among model families. Hierarchical architectures show a second, more dataset-dependent pattern: Swin~\citep{Liu2021Swin}, PVT~\citep{Wang2021PVT}, and CoAtNet~\citep{dai2021coatnet} exceed isotropic ViT~\citep{dosovitskiy2021vit} and DeiT~\citep{touvron21a} on Flowers-102 ($p<0.001$), while the difference does not reach significance on the other five datasets. Rank order is nevertheless fairly consistent across related datasets, including $\rho=0.95$ between CIFAR-10 and CIFAR-100, suggesting that model family contributes to stability beyond any one dataset.

\subsection{Geometric Stability is Not Effective Dimensionality}
\label{sec:not_eff_dim}
One possible explanation is that Shesha\textsubscript{FS} is largely a measure of effective dimensionality, where a representation that spreads variance across many dimensions might be expected to leave both random feature halves informative. The benchmark does not support that explanation. Across 170 models and six datasets, Shesha\textsubscript{FS} is negatively correlated with participation ratio (\citealp{Gao2017}; mean Spearman $\rho=-0.36$; all six datasets significant at $p<0.01$, ranging from $-0.29$ on CIFAR-10 to $-0.47$ on EuroSAT) and positively correlated with top-eigenvalue variance share (mean $\rho=+0.39$). In these models, higher effective dimensionality is therefore not associated with higher feature-split stability.

DINOv2 makes the distinction especially clear. On the natural-image datasets it has among the lowest Shesha\textsubscript{FS} values in the benchmark (rank 36/36 on CIFAR-10) but the highest participation ratio among the 36 families (98.98 on CIFAR-10, compared with a benchmark mean of 51.64; 227.85 on CIFAR-100). Within DINOv2, the scale trend can also reverse. On CIFAR-10, the giant model has a lower participation ratio than the large model (87.9 vs.\ 117.7) while having higher Shesha\textsubscript{FS}. Effective dimensionality therefore does not determine feature-split stability.

This difference follows from the quantities themselves. Participation ratio depends only on the eigenvalue spectrum and is invariant to rotations of the feature basis. Shesha\textsubscript{FS}, by contrast, depends on how the representation is expressed across coordinates. Two representations can therefore share the same spectrum while differing in feature-split stability. The PCA-coordinate analysis in Sec.~\ref{sec:mechanism} provides a controlled example of this basis dependence. In the empirical PCA sweeps, Shesha\textsubscript{FS} decreases as the representation is expressed in progressively fewer principal-component coordinates, but this is an observed effect of that transformation, not a general identity between stability and dimensionality. Geometric stability and effective dimensionality are distinct quantities, and they are negatively associated in this benchmark.

\subsection{Geometric Stability is Distinct From Corruption Robustness}
\label{sec:vision-corrupt}
We next asked whether a representation's clean geometric stability anticipates its robustness to distribution shift. For each model we computed $\Delta\mathrm{LogME}$, the drop in LogME from the clean to the corrupted evaluation (CIFAR-10-C and CIFAR-100-C~\citep{hendrycks2019robustness}, severity 5, 19 corruption types), and related it to clean
Shesha\textsubscript{FS} by partial Spearman correlation, controlling for clean LogME so that the relationship is not driven by high-transfer models simply having more to lose ($N = 170$ models per dataset; bootstrap 95\% confidence intervals, 10{,}000 resamples).

The predicted relationship does not appear consistently. If higher geometric stability implied greater corruption robustness, more stable models would show smaller LogME drops and the correlation would be negative. On CIFAR-100 there is no clear relationship ($\rho_{\textrm{partial}}=-0.03$, 95\% CI $[-0.20,+0.27]$ across all 19 corruptions). On CIFAR-10 the aggregate correlation is weakly positive ($\rho_{\textrm{partial}}=+0.26$, 95\% CI $[+0.08,+0.43]$), opposite to that prediction. Because the effect is small and does not replicate on CIFAR-100, we interpret the two datasets as showing no consistent association between clean feature-split stability and corruption robustness.

Together with the subset-reliability result in Sec.~\ref{sec:probe_sensitivity}, this helps delimit what Shesha\textsubscript{FS} measures. It is associated with the consistency of analyses across feature subsets, but not with robustness to input corruption in these experiments. Shesha\textsubscript{FS} should therefore be read as a diagnostic of feature-basis redundancy rather than a general measure of representation quality. 

\begin{table}[H]
\caption{DINO generations across six datasets ($N=36$ families per dataset).
For each dataset and generation, we report family-mean LogME,
Shesha\textsubscript{FS}, linear-probe variability across $K=100$ random
feature halves ($\sigma_{\mathrm{probe}}$), and participation ratio (PR),
together with rank among the 36 families. For LogME, Shesha\textsubscript{FS},
and PR, rank 1 denotes the highest family mean; for
$\sigma_{\mathrm{probe}}$, rank 1 denotes the lowest probe variability.
DINOv2 combines high transferability with low geometric stability on several
datasets, while often retaining low probe variability, illustrating that
task alignment a feature-split geometric stability capture distinct
components of subset-probe reliability.}
\label{tab:dino}
\centering
\scriptsize
\resizebox{\textwidth}{!}{%
\begin{tabular}{lc cc cc cc cc}
\hline
Dataset &
DINO &
LogME &
LogME Rank &
Shesha\textsubscript{FS} &
$\mathcal{S}$ Rank &
$\sigma_{\mathrm{probe}}$ &
Probe Rank &
PR &
PR Rank \\
\hline

\multirow{3}{*}{CIFAR-10}
 & v1 & 0.425 & 24/36 & 0.639 & 27/36 & 0.0048 & 31/36 & 54.94 & 20/36 \\
 & v2 & 1.013 &  2/36 & 0.368 & 36/36 & 0.0019 &  4/36 & 98.98 &  1/36 \\
 & v3 & 0.819 &  9/36 & 0.691 & 19/36 & 0.0022 &  8/36 & 40.23 & 30/36 \\
\hline

\multirow{3}{*}{CIFAR-100}
 & v1 & 1.045 & 28/36 & 0.579 & 28/36 & 0.0061 & 33/36 &  90.08 & 14/36 \\
 & v2 & 1.373 &  2/36 & 0.268 & 35/36 & 0.0034 &  4/36 & 227.85 &  1/36 \\
 & v3 & 1.206 &  9/36 & 0.588 & 27/36 & 0.0044 & 10/36 &  82.52 & 20/36 \\
\hline

\multirow{3}{*}{Flowers-102}
 & v1 & 1.260 & 21/36 & 0.852 & 19/36 & 0.0028 & 13/36 & 39.45 & 15/36 \\
 & v2 & 2.586 &  1/36 & 0.319 & 36/36 & 0.0005 &  1/36 & 60.10 &  4/36 \\
 & v3 & 1.709 &  9/36 & 0.766 & 27/36 & 0.0010 &  5/36 & 39.40 & 16/36 \\
\hline

\multirow{3}{*}{DTD}
 & v1 & 0.692 & 25/36 & 0.448 & 32/36 & 0.0102 & 33/36 &  76.31 & 26/36 \\
 & v2 & 0.876 & 13/36 & 0.474 & 29/36 & 0.0075 &  5/36 & 104.51 & 16/36 \\
 & v3 & 0.814 & 15/36 & 0.602 & 17/36 & 0.0068 &  1/36 &  67.89 & 30/36 \\
\hline

\multirow{3}{*}{EuroSAT}
 & v1 & 0.547 & 10/36 & 0.775 & 35/36 & 0.0035 & 22/36 & 24.01 & 20/36 \\
 & v2 & 0.568 &  6/36 & 0.951 &  3/36 & 0.0030 & 11/36 & 16.09 & 26/36 \\
 & v3 & 0.588 &  4/36 & 0.924 &  8/36 & 0.0025 &  3/36 & 13.63 & 29/36 \\
\hline

\multirow{3}{*}{Oxford Pets}
 & v1 & 0.755 & 31/36 & 0.771 & 11/36 & 0.0073 & 30/36 & 23.28 & 32/36 \\
 & v2 & 1.301 & 12/36 & 0.533 & 33/36 & 0.0043 &  3/36 & 42.61 & 18/36 \\
 & v3 & 0.717 & 33/36 & 0.777 &  8/36 & 0.0080 & 33/36 & 28.71 & 31/36 \\
\hline
\end{tabular}
}
\end{table}

\subsection{Geometric Stability and Probe Subset-Sensitivity}
\label{sec:probe_sensitivity}

\begin{figure}[h]
\centering
\includegraphics[width=\textwidth]{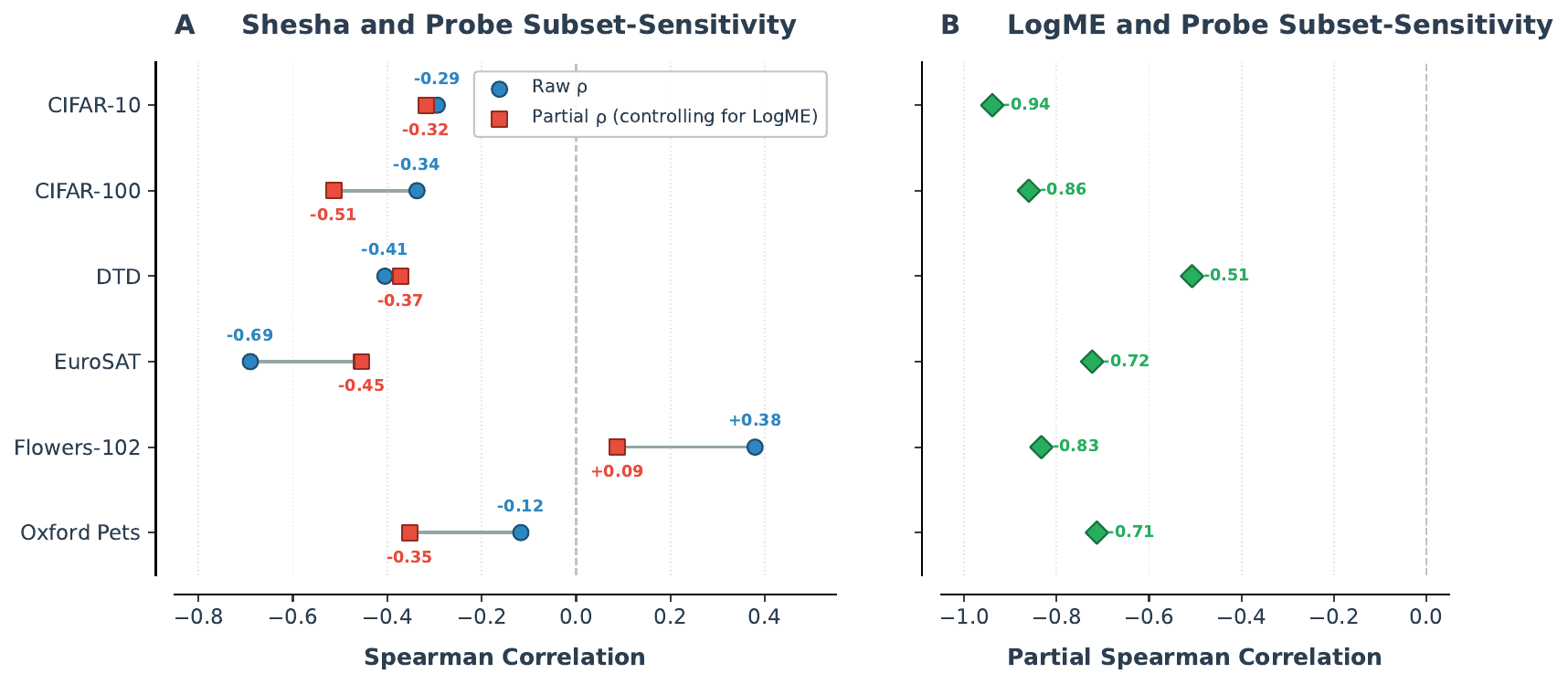}
\caption{Geometric stability, task alignment, and probe
subset-sensitivity across six vision datasets.
For each dataset, probe variability
($\sigma_{\mathrm{probe}}$) is the standard deviation of linear-probe test
accuracy across $K=100$ random feature halves for 170 models.
A. Raw Spearman correlations between
Shesha\textsubscript{FS} and $\sigma_{\mathrm{probe}}$, together with partial
correlations controlling for LogME. After accounting for LogME, higher
Shesha\textsubscript{FS} is associated with lower probe variability on five
of six datasets; Flowers-102 shows no significant conditional association.
B. Partial Spearman correlations between LogME and
$\sigma_{\mathrm{probe}}$ controlling for Shesha\textsubscript{FS}.
LogME remains negatively associated with probe variability on all six
datasets. Together, the results indicate that task alignment is the more
consistent correlate of subset-probe reliability, while feature-split
geometric stability provides complementary information about sensitivity to
feature selection.}
\label{fig:probe_sensitivity}
\end{figure}

A more direct test of the practical relevance of feature-split stability is whether it is related to sensitivity to the choice of feature subset. We tested this using linear probes, where the subset of representation coordinates can be varied while holding the examples, labels, and train/test split fixed.

For each of 170 vision models, we extracted representations on six clean image datasets: CIFAR-10, CIFAR-100, Flowers-102, DTD, EuroSAT, and Oxford Pets. For each model and dataset, we trained logistic-regression probes on $K=100$ random halves of the feature dimensions and measured the standard deviation of probe test accuracy across subsets, denoted $\sigma_{\mathrm{probe}}$. If feature-split geometric stability is related to the reliability of subset-based analyses, higher Shesha\textsubscript{FS} should generally be associated with lower $\sigma_{\mathrm{probe}}$.

The raw association was heterogeneous across datasets. Shesha\textsubscript{FS} was negatively correlated with probe variability on CIFAR-10 ($\rho=-0.294$), CIFAR-100 ($\rho=-0.337$), DTD ($\rho=-0.405$), and EuroSAT ($\rho=-0.690$), whereas the association was weak on Oxford Pets ($\rho=-0.117$) and positive on Flowers-102 ($\rho=+0.379$). This heterogeneity indicates that feature-split stability alone does not determine the reliability of downstream probes.

Probe variability was also strongly related to task alignment, as measured by LogME. We therefore computed partial Spearman correlations between Shesha\textsubscript{FS} and $\sigma_{\mathrm{probe}}$ while controlling for LogME. Conditional on LogME, higher Shesha\textsubscript{FS} was associated with lower probe variability on five of six datasets: CIFAR-10 ($\rho_{\mathrm{partial}}=-0.318$, $p=2.49\times10^{-5}$), CIFAR-100 ($-0.513$, $p=1.03\times10^{-12}$), DTD ($-0.371$, $p=6.81\times10^{-7}$), EuroSAT ($-0.454$, $p=5.49\times10^{-10}$), and Oxford Pets ($-0.352$, $p=2.70\times10^{-6}$). Flowers-102 showed no significant conditional association ($\rho_{\mathrm{partial}}=+0.087$, $p=0.263$). Conversely, LogME remained strongly negatively associated with probe variability after controlling for Shesha\textsubscript{FS} on all six datasets, with partial correlations ranging from $-0.507$ to $-0.939$.

These results suggest that task alignment is the more consistent correlate of subset-probe reliability, while feature-split geometric stability provides complementary information about sensitivity to feature selection. This distinction is particularly clear on Flowers-102, where the positive marginal association between Shesha\textsubscript{FS} and probe variability largely disappears after accounting for task alignment. As secondary controls, the Shesha--variability association also remained negative after controlling for mean probe accuracy on CIFAR-10, CIFAR-100, DTD, and EuroSAT, while it was nonsignificant on Flowers-102 and Oxford Pets. These analyses therefore support Shesha\textsubscript{FS} as a diagnostic of one component of subset sensitivity, rather than as a universal predictor of downstream probe robustness.

\section{Discussion}
%%%%%%%%%%%%%%%%%%%%%%%%%%%%%%%%%%%%%%%%%%%%%%%%%%%%%%%%%%%%%%%%%%%%%%%%%%%
\label{sec:discussion}
Geometric stability measures how consistently the sample geometry of a representation is recovered from complementary subsets of its feature coordinates. This is different from representational similarity. Similarity measures compare two representations, often after quotienting out changes of basis; Shesha asks how much the geometry of one representation depends on which coordinates are observed. Across the experiments here, that distinction is consequential but not uniformly beneficial: stability varies with transformation regime, can be low in high-performing encoders, and predicts the variability of linear probes trained on random feature subsets. These results motivate treating stability as a separate descriptive property when an analysis depends on feature subsets, not as a replacement for similarity or a general measure of representation quality.

\subsection{Similarity and Feature-Split Stability}

RSA~\citep{Kriegeskorte2008} and CKA~\citep{kornblith2019similarity} compare the geometry of two representations while abstracting away from direct feature correspondence. Related methods, including SVCCA~\citep{raghu2017svcca}, PWCCA~\citep{morcos2018insights}, and Procrustes analysis~\citep{Schnemann1966,Rohlf1990,Masarotto2018,dryden1998statistical}, make different choices about which transformations should be treated as irrelevant. These methods answer a comparison question. Shesha instead measures agreement between RDMs obtained from complementary feature subsets of the same representation.

The distinction is formal. For an orthogonal matrix $Q$,
\[
(XQ)(XQ)^\top = XX^\top,
\]
so rotation-invariant similarity measures cannot distinguish $X$ from $XQ$. Shesha can, because selecting coordinates before and after the rotation generally produces different subspaces. The counterexample in Appendix~\ref{si:proofs} makes this explicit: an orthogonal rotation leaves the full Gram matrix unchanged while changing $\mathcal{S}$. Basis dependence is therefore part of the quantity being measured, not an approximation error.

This also clarifies what the spectral experiments establish. CKA between an original and modified representation measures similarity to a reference representation. Shesha of the modified representation measures agreement between its feature subsets. The two quantities need not respond in the same way to spectral deletion or compression because they have different estimands. The results in Sec.~\ref{sec:framework} show that this difference is systematic across transformation regimes; they do not imply that one statistic recovers geometry that the other fails to measure.

Recent work has emphasized that representational similarity scores themselves require careful interpretation. CKA can change without corresponding changes in model behavior~\citep{davari2023reliability}, and biased estimators can behave poorly in low-sample, high-dimensional settings~\citep{murphy2024correcting}. Conversely, CKA and CCA admit useful functional interpretations in terms of linear readout alignment~\citep{Harvey2024}. \cite{williams2024equivalence} further showed close connections between centered RSA and CKA. These results help delimit what similarity measures mean. Shesha addresses a separate question and should be interpreted on those terms.

An obvious concern is that a basis-dependent statistic can depend on an arbitrary coordinate system. That concern is real. Shesha is most useful when the native coordinates are themselves part of the analysis, as in feature subsetting, pruning, dropout, or probes restricted to subsets of units. In settings where only basis-invariant properties matter, rotation sensitivity is not an advantage. The whitening analysis makes the same point empirically: changing the spectrum and coordinate structure changes the relationship between Shesha and similarity, so the practical value of Shesha depends on the representation and on the operation being studied.

The analogy to reliability measures is useful but limited. Split-half and noise-ceiling approaches in RSA assess the reproducibility of a representational geometry across observations or measurements~\citep{Nili2014,Walther2016}; each RDM is still computed from the full available feature population. Shesha instead partitions the feature population itself, so each half-RDM is computed from only a subset of the representation's coordinates. Both therefore assess reproducibility under a specified partition, but they perturb different objects and have different statistical interpretations. In particular, $\mathcal{S}$ is not a noise ceiling and does not bound the attainable similarity between representations.

\subsection{Geometric Stability as a Distinct Selection Axis}
\label{sec:discussion-distinct-axis}

The vision benchmark shows why this distinction can matter in model comparison. Transfer performance and geometric stability are only weakly related across architectural families ($\rho=+0.21$, not significant), so high transfer does not imply either high or low stability. DINOv2 is a clear example: it combines strong transfer performance with bottom-quartile stability on five of the six datasets, despite having the highest participation ratio in the benchmark. Its low Shesha score therefore cannot be reduced to a simple low-dimensionality account.

This does not imply a general trade-off between transfer and stability. CLIP- and SigLIP-family models often combine high transfer with high stability, and the family-level trend is not negative. We use the term \emph{geometric tax} descriptively for cases in which strong transfer is accompanied by comparatively low feature-split stability. The benchmark shows that such a tax can occur, not that it is an inevitable consequence of transfer learning or self-supervision.

DINOv3 is informative for the same reason, but the comparison should not be read causally. DINOv3 has substantially higher stability than DINOv2 on the benchmark while retaining strong transfer performance. Its training recipe includes Gram anchoring, among other changes. The result is consistent with the possibility that training choices affecting coordinate structure can alter stability, but the generational comparison does not isolate Gram anchoring as the cause. A controlled objective or regularizer ablation on a fixed architecture would be needed for that claim.

The same caution applies to the broader family comparisons. Contrastively trained models are generally more stable than the single-modality self-supervised models in this benchmark, and EVA-02 is highly stable despite using a reconstruction-style objective. These associations suggest that stability depends on more than architecture or scale alone, but they do not identify a unique training mechanism.

For model selection, the practical implication is conditional. If the downstream use only requires full-representation prediction, transfer performance may be the relevant criterion. If the planned analysis repeatedly selects, removes, or samples feature coordinates, then feature-split stability provides additional information that transfer metrics do not report. The corruption experiments also place a boundary on this interpretation: Shesha does not show a consistent relationship with input-level corruption robustness, so it should not be treated as a general robustness score.

\subsection{Feature-Subset Analyses and Interpretability}
\label{sec:discussion-mech-interp}

The direct functional evidence in this paper concerns linear probes. Linear probing is widely used to test whether information is accessible from a representation~\citep{Alain2016UnderstandingIL}. Across six vision datasets, the marginal relationship between Shesha\textsubscript{FS} and variability in probe accuracy across random feature subsets is heterogeneous (Sec.~\ref{sec:probe_sensitivity}), which tells us that probe subset-sensitivity is not determined by feature-split geometric stability alone.

A major additional factor is task alignment. Recent theoretical work on pretraining and linear probing emphasizes that downstream generalization depends not only on representation dimensionality, but also on how well the learned representation aligns with the downstream task~\citep{njaradi2026optimal}. Consistent with this view, LogME is strongly negatively associated with probe variability across all six datasets, including after controlling for Shesha\textsubscript{FS}. Conditional on LogME, however, higher Shesha\textsubscript{FS} remains associated with lower probe variability on five of six datasets, with Flowers-102 showing no significant conditional association. Secondary controls for mean probe accuracy support the same broad pattern. These results suggest that task alignment is the more consistent correlate of subset-probe reliability, while feature-split geometric stability contributes complementary information about sensitivity to the particular coordinates available to the probe.

This distinction also clarifies how Shesha\textsubscript{FS} should be interpreted. A low score means that complementary feature subsets induce different pairwise geometries, but it does not imply that every downstream analysis will be unstable. A representation can have low Shesha\textsubscript{FS} while supporting highly consistent probes when the available features are strongly aligned with the downstream task. Conversely, a high Shesha\textsubscript{FS} indicates greater agreement across sampled feature partitions, but does not by itself guarantee strong or invariant downstream performance. Shesha\textsubscript{FS} therefore characterizes one geometric source of subset sensitivity rather than serving as a universal measure of interpretability or probe robustness.

Other interpretability methods may also depend on the native coordinate basis, but they are not tested here. Activation patching, steering, sparse feature decompositions, and related interventions introduce additional assumptions about causal structure, feature identity, or intervention semantics that are not captured by feature-split consistency. Their relationship to Shesha\textsubscript{FS} should therefore be established directly rather than inferred from the probe results. We restrict the empirical interpretability claim in this paper to sensitivity of linear probes to feature-subset choice.

\subsection{Geometric Stability Across Substrates}

The same statistic can be computed for any representation matrix whose columns define a meaningful feature basis. We therefore applied it not only to learned encoders but also to protein representations~\citep{uniprot2023uniprot}, molecular profiles~\citep{Zheng2017}, and neural population recordings~\citep{steinmetz2019distributed}. The purpose of these analyses is not to argue for a common mechanism across biological and artificial systems. Rather, they test whether the distinction between similarity and feature-split stability is specific to deep-network activations.

Our results show that biological representations exhibit qualitatively similar relationships between similarity and feature-split stability to those observed in machine learning representations.  In the protein experiments, PCA-coordinate compression produces the same qualitative dissociation between similarity and stability seen in the controlled compression regime. In the neural recordings, the relationship between the two quantities is weak, as it is for several learned encoder families without explicit compression. These results show that feature-split stability is measurable outside artificial neural networks and can vary independently of similarity in those settings as well.

The interpretation of a feature split is domain-specific. For a neural population, features are recorded units; for molecular data, they may be genes or measured variables; for a learned encoder, they are model coordinates. A low Shesha value therefore does not imply the same biological or computational mechanism in each case. It means only that complementary subsets of the measured features induce different sample geometries.

This distinction is also important when comparing Shesha with established reliability analyses in neuroscience. RSA noise ceilings and split-half reliability quantify variability across observations, subjects, or repeated measurements while each RDM is computed from the full available feature population~\citep{Nili2014,Walther2016}. Shesha instead partitions the feature population itself, so each half-RDM is constructed from only a subset of the representation's coordinates. The two procedures therefore probe different sources of variability and have different statistical interpretations; neither should be interpreted as a bound on the other.

\subsection{Limitations}
\label{sec:discussion-limitations}

Shesha is a global summary. It reduces feature-split agreement for an entire representation to a single scalar and can therefore hide localized instability affecting particular categories, stimuli, or subspaces. Per-class or local variants could address this, but they are not developed here.

The metric is also intentionally basis-dependent. This makes it appropriate for questions involving the native coordinates of a representation, but it also means that values cannot be interpreted as intrinsic properties of the represented manifold. Orthogonally equivalent representations can have different Shesha values by construction. That distinction should be kept in mind when comparing models whose coordinate systems have different meanings.

The theoretical results establish non-invariance, not a universal direction of change. In particular, the PCA result in Appendix~\ref{si:proofs} shows that PCA-coordinate compression can change Shesha; it does not prove that every PCA compression must reduce it. The monotonic decreases observed across the empirical compression experiments are therefore empirical findings rather than consequences of the theorem.

Feature extraction for the vision benchmark uses a single seed per dataset, except for CIFAR-10, which we recomputed across three seeds for all 170 models with near-identical rankings (Spearman $\rho \geq 0.993$, median per-model CV 0.75\%; Appendix~\ref{si:seed}). The remaining five datasets should be read as point estimates. Cross-domain stability estimates average over 15 seeds and are not subject to this caveat.

The encoder transformation framework covers a deliberately small set of controlled perturbations. It does not characterize all changes encountered under fine-tuning, distribution shift, quantization, pruning, or other post-training procedures. Likewise, the subset-probe analysis tests one concrete consequence of feature-split stability. We do not test activation patching or steering in this study, and no claim about those interventions follows directly from the present results.

The cross-domain analyses also differ substantially in sample size, feature semantics, and data-generating process. The neuroscience result ($N=846$ configurations from 26 sessions) comes from a specific recording paradigm and should not be assumed to generalize across brain regions, species, or behavioral contexts. The biological experiments establish applicability of the statistic, not a substrate-independent mechanism.

Finally, the benchmark does not identify the cause of the family-level differences in stability. DINOv2 and DINOv3 differ in multiple aspects of training, and contrastive and self-supervised families differ in architecture, data, objective, and optimization. The SAM and supervision ablations provide controlled examples in which training changes alter stability, but they do not isolate the mechanism behind the DINO-family differences. Establishing causal effects of particular objectives or regularizers will require matched training experiments.

\subsection{Outlook}
\label{sec:discussion-outlook}

The most immediate use of geometric stability is as a companion diagnostic for analyses that depend on feature subsets. The \href{https://pypi.org/project/shesha-geometry/}{\texttt{shesha-geometry}} package~\citep{shesha2026} takes a representation matrix as input and estimates $\mathcal{S}$ without labels, repeated measurements, or a reference representation. This makes it straightforward to report alongside similarity or transfer metrics when coordinate-level reliability is relevant.

Several questions remain open. It is not yet clear which training choices causally control stability, whether stability can be increased without changing downstream behavior, or when higher feature redundancy is desirable rather than wasteful. The DINOv2--DINOv3 comparison and the controlled SAM results suggest that stability is trainable, but they do not identify a unique route. More direct interventions on the training objective or regularization would test that possibility.

More broadly, the results here argue for keeping three questions separate: whether two representations are similar, whether a representation supports a downstream task, and whether its geometry is reproducible across feature subsets. In some models these properties coincide; in others they do not. Shesha is intended to measure the third.

\section*{Declaration of Generative AI Usage}
Large language models (specifically the GPT, Claude, Gemini, and Grok families) were used to assist with (limited) code factoring, code review and debugging, and text reviewing, editing, and polishing. All hypotheses, experimental designs, analyses, and interpretations were independently formulated and verified by the author, and the author assumes full responsibility for all content and claims in this work.


% \newpage

% \newpage
% \vskip 0.2in
% \bibliography{sample}

% \newpage
\appendix

% \section{}

% Note: in this sample, the section number is hard-coded in. Following
% proper LaTeX conventions, it should properly be coded as a reference:

%In this appendix we prove the following theorem from
%Section~\ref{sec:textree-generalization}:

% 1. Restart the figure counter
\setcounter{figure}{0}

% 2. Keep the label text strictly as "Figure"
\renewcommand{\figurename}{Figure}

% 3. Add the "S" directly to the number format with NO space
\renewcommand{\thefigure}{S\arabic{figure}}

% 1. Restart the table counter
\setcounter{table}{0}

% 2. Keep the label text strictly as "Table"
\renewcommand{\tablename}{Table}

% 3. Add the "S" directly to the number format with NO space
\renewcommand{\thetable}{S\arabic{table}}

% \section{}

\section{Shesha Variants}
\label{si:variants}

The main text presents Feature-Split Shesha (Shesha\textsubscript{FS}), the variant analyzed throughout this paper. The split-half construction can in principle be adapted to other partitioning schemes, but the formal results and empirical claims here concern Shesha\textsubscript{FS} only.

\subsection{Feature-Split Shesha (Shesha\texorpdfstring{\textsubscript{FS}}{FS})}
The primary variant, described in the main text. Feature dimensions $\{1,\ldots,d\}$ are randomly partitioned into two disjoint halves $F^{(1)}_k, F^{(2)}_k$; an RDM is computed from each half using cosine distance; and the Spearman rank correlation between the two vectorized upper triangles is computed for each partition. The population quantity $\mathcal{S}(X)$ is the expectation over balanced complementary partitions, and the experiments use the Monte Carlo estimator $\widehat{\mathcal{S}}_{K}(X)$ with $K{=}30$. This variant measures agreement of sample geometry across complementary subsets of the feature basis and requires no labels or repeated measurements.

\subsection{Sample-Split Shesha (Shesha\texorpdfstring{\textsubscript{SS}}{SS})}
A sample-split analogue requires correspondence between the two halves, such as repeated observations of the same conditions. Under that design, repeated observations are divided into two groups, one condition-level representation is estimated from each group, and the two RDMs over the common condition set are correlated. This probes reproducibility across observations rather than across feature coordinates. Sample-Split Shesha is not used or evaluated in the present paper and is included only to distinguish this possible observation-axis construction from Shesha\textsubscript{FS}.

\subsection{Anchor Shesha (Shesha\texorpdfstring{\textsubscript{anc}}{anc})}
Anchor-based variants replace the full $O(n^2)$ RDM with distances to a fixed set of $m$ reference samples, reducing the number of pairwise distances to $O(nm)$. Such variants require a specified correspondence or aggregation rule for comparing anchor-distance profiles across splits, and their behavior depends on the number and choice of anchors. We do not analyze an anchor-based estimator in the present paper; all reported results use the full-RDM Shesha\textsubscript{FS} definition.

\section{Invariance Proofs and Counterexamples}
\label{si:proofs}

We prove the exact invariance and non-invariance properties listed in Sec.~\ref{sec:invariances} of the main text. Throughout, $X\in\mathbb{R}^{n\times d}$ is a representation matrix and $\Pi_d$ is the uniform distribution over balanced complementary feature partitions $(A,B)$, as defined in Eq.~\eqref{eq:shesha_full}. We distinguish the population functional
\[
\mathcal{S}(X)
=
\mathbb{E}_{(A,B)\sim\Pi_d}
\left[
\rho_s\!\left(
\operatorname{vec}_{\triangle}D(X_A),
\operatorname{vec}_{\triangle}D(X_B)
\right)
\right]
\]
from its finite-$K$ Monte Carlo estimator $\widehat{\mathcal{S}}_K(X)$ in Eq.~\eqref{eq:shesha_estimator}. Unless stated otherwise, invariance claims below concern the population functional $\mathcal{S}$. We assume that the cosine distances and Spearman correlations appearing in the relevant partitions are well defined, i.e., the restricted sample vectors have nonzero norm and the RDM rank vectors are nonconstant.

\subsection{Uniform Scalar Scaling Invariance}
\label{app:proof-scaling}
\begin{proof}
Let $Y=\alpha X$ for any nonzero scalar $\alpha$. For either half $s\in\{A,B\}$ and any pair of samples $i,j$,
\[
1-
\frac{(\alpha x_i^{(s)})\cdot(\alpha x_j^{(s)})}
{\|\alpha x_i^{(s)}\|\,\|\alpha x_j^{(s)}\|}
=
1-
\frac{\alpha^2 x_i^{(s)}\cdot x_j^{(s)}}
{|\alpha|^2\|x_i^{(s)}\|\,\|x_j^{(s)}\|}
=
1-
\frac{x_i^{(s)}\cdot x_j^{(s)}}
{\|x_i^{(s)}\|\,\|x_j^{(s)}\|}.
\]
Thus $D(Y_s)=D(X_s)$ for every balanced partition and both halves. The split-half Spearman correlation is therefore unchanged pointwise in the partition, and hence
\[
\mathcal{S}(\alpha X)=\mathcal{S}(X).
\]
The same equality holds exactly for $\widehat{\mathcal{S}}_K$ when evaluated using the same sampled partitions.
\end{proof}

\subsection{Feature Permutation Invariance}
\label{app:proof-permutation}
\begin{proof}
Let $Y=XP$, where $P$ is a $d\times d$ permutation matrix. Define the induced permutation $p$ of feature indices by
\[
Y_{:,j}=X_{:,p^{-1}(j)}.
\]
For a balanced partition $\pi=(A,B)$, define
\[
p^{-1}\pi=\bigl(p^{-1}(A),p^{-1}(B)\bigr).
\]
Restricting $Y$ to coordinates in $A$ selects the same original coordinates as restricting $X$ to $p^{-1}(A)$, up to a permutation within the selected coordinates. Coordinate permutation preserves inner products and norms, so
\[
D(Y_A)=D\!\left(X_{p^{-1}(A)}\right),
\qquad
D(Y_B)=D\!\left(X_{p^{-1}(B)}\right).
\]
Therefore the split statistic satisfies
\[
\rho_{\pi}(Y)=\rho_{p^{-1}\pi}(X),
\]
where $\rho_{\pi}$ denotes the Spearman correlation between the two half-RDMs for partition $\pi$.

Because $\Pi_d$ is uniform over balanced partitions, it is invariant under relabeling of feature indices. Hence, if $\pi\sim\Pi_d$, then $p^{-1}\pi\sim\Pi_d$, and
\[
\begin{aligned}
\mathcal{S}(XP)
&=
\mathbb{E}_{\pi\sim\Pi_d}\!\left[\rho_{\pi}(XP)\right] \\
&=
\mathbb{E}_{\pi\sim\Pi_d}\!\left[\rho_{p^{-1}\pi}(X)\right] \\
&=
\mathbb{E}_{\pi'\sim\Pi_d}\!\left[\rho_{\pi'}(X)\right]
=
\mathcal{S}(X).
\end{aligned}
\]
For independently sampled Monte Carlo partitions, $\widehat{\mathcal{S}}_K(XP)\overset{d}{=}\widehat{\mathcal{S}}_K(X)$. If the sampled partitions are coupled by the same permutation, the finite-$K$ estimates are equal exactly.
\end{proof}

\subsection{Non-Invariance to PCA Coordinate Compression}
\label{si:proofs-pca}

Here PCA compression means retaining the leading PCA score coordinates, $T_r(X)=XV_r\in\mathbb{R}^{n\times r}$ (for centered $X$), rather than forming the rank-$r$ reconstruction $XV_rV_r^\top$ back in the original $d$-dimensional coordinate basis. We establish non-invariance by explicit counterexample.

\begin{proof}
Let
\[
u=
\begin{pmatrix}
1\\
1\\
-1\\
-1
\end{pmatrix},
\qquad
v=
\begin{pmatrix}
1\\
-1\\
1\\
-1
\end{pmatrix},
\qquad
u^\top v=0,
\]
and define
\[
X=
\begin{pmatrix}
2 & 1 & 2 & 1\\
2 & -1 & 2 & -1\\
-2 & 1 & -2 & 1\\
-2 & -1 & -2 & -1
\end{pmatrix}
=
\begin{pmatrix}
2u & v & 2u & v
\end{pmatrix}.
\]
The columns of $X$ are centered. For $d=4$, the three complementary bipartitions of the coordinates are $\{1,2\}\mid\{3,4\}$, $\{1,3\}\mid\{2,4\}$, and $\{1,4\}\mid\{2,3\}$. (If $\Pi_4$ is represented using ordered halves, each bipartition appears twice and the expectation is unchanged.)

For $\{1,2\}\mid\{3,4\}$ the two halves are identical, so their RDM correlation is $1$. For $\{1,4\}\mid\{2,3\}$ the halves differ only by a coordinate permutation, so their RDM correlation is also $1$. For $\{1,3\}\mid\{2,4\}$, one half contains two copies of $u$ and the other two copies of $v$. In the pair order $(12,13,14,23,24,34)$, the corresponding one-dimensional cosine-distance vectors are proportional to
\[
(0,2,2,2,2,0)
\quad\text{and}\quad
(2,0,2,2,0,2),
\]
whose Spearman correlation is $-1/2$. Therefore
\[
\mathcal{S}(X)
=
\frac{1}{3}\left(1-\frac12+1\right)
=
\frac12.
\]

Now compute the PCA score representation. Since
\[
X^\top X
=
\begin{pmatrix}
16&0&16&0\\
0&4&0&4\\
16&0&16&0\\
0&4&0&4
\end{pmatrix},
\]
its two nonzero principal directions can be chosen as
\[
q_1=\frac{1}{\sqrt2}(1,0,1,0)^\top,
\qquad
q_2=\frac{1}{\sqrt2}(0,1,0,1)^\top.
\]
Thus the rank-$2$ PCA score representation is
\[
Z=XV_2
=
\begin{pmatrix}
Xq_1 & Xq_2
\end{pmatrix}
=
\begin{pmatrix}
2\sqrt2\,u & \sqrt2\,v
\end{pmatrix}.
\]
With two coordinates, the only balanced split compares the first PCA coordinate with the second. Uniform scaling does not change either one-dimensional cosine RDM, so
\[
\mathcal{S}(Z)
=
\rho_s\!\left(D(u),D(v)\right)
=
-\frac12.
\]
Hence
\[
\mathcal{S}(XV_2)=-\frac12
\neq
\frac12=\mathcal{S}(X),
\]
which proves that Shesha\textsubscript{FS} is not invariant to PCA coordinate compression.

In this example $\operatorname{rank}(X)=2$, so the retained principal components span the entire nonzero row space of $X$. Consequently
\[
ZZ^\top
=
XV_2V_2^\top X^\top
=
XX^\top.
\]
Hence the rowwise inner products and norms are also identical, so the full cosine-distance RDM is the same for $X$ and $Z$. Thus linear $\operatorname{CKA}(X,Z)=1$ exactly, and an RSA comparison based on these full cosine-distance RDMs likewise cannot distinguish the two representations, while $\mathcal{S}(Z)\neq\mathcal{S}(X)$.

The example is deliberately lossless, isolating the change of coordinates from any loss of information: $Z$ is obtained by an orthogonal rotation into the principal-coordinate basis followed by removal of zero-variance coordinates.
\end{proof}

\paragraph{Scope.}
The counterexample establishes non-invariance: PCA coordinate compression can change $\mathcal{S}$. It does not imply that every PCA compression changes $\mathcal{S}$, or changes it in a particular direction. The result concerns PCA score coordinates; a rank-$r$ reconstruction expressed in the original $d$-dimensional basis is a different object about which we make no claim.

\subsection{Non-Invariance to Orthogonal Transformations: Constructive Counterexample}
\label{si:proof-ortho-transform}
\begin{proof}
We use the same representation
\[
X=
\begin{pmatrix}
2 & 1 & 2 & 1\\
2 & -1 & 2 & -1\\
-2 & 1 & -2 & 1\\
-2 & -1 & -2 & -1
\end{pmatrix},
\]
for which the preceding proof established $\mathcal{S}(X)=1/2$. Let
\[
Q=
\begin{pmatrix}
\frac35 & 0 & \frac45 & 0\\
0 & 1 & 0 & 0\\
-\frac45 & 0 & \frac35 & 0\\
0 & 0 & 0 & 1
\end{pmatrix}.
\]
The $2\times2$ block acting on coordinates $1$ and $3$ is a rotation, so
$Q^\top Q=I$. Setting $Y=XQ$ gives
\[
Y=
\begin{pmatrix}
-\frac25 & 1 & \frac{14}{5} & 1\\
-\frac25 & -1 & \frac{14}{5} & -1\\
\frac25 & 1 & -\frac{14}{5} & 1\\
\frac25 & -1 & -\frac{14}{5} & -1
\end{pmatrix}.
\]
No row is zero on either half of any balanced split, so all cosine RDMs in this counterexample are well defined.

Again consider the three complementary bipartitions. For the split $\{1,3\}\mid\{2,4\}$, the first half depends only on the sign pattern $u$ and the second only on $v$, giving the same Spearman correlation $-1/2$ as above. For either mixed split, $\{1,2\}\mid\{3,4\}$ or $\{1,4\}\mid\{2,3\}$, the two upper-triangular RDM rank vectors, in pair order $(12,13,14,23,24,34)$, are
\[
\left(\frac72,\frac32,\frac{11}{2},\frac{11}{2},\frac32,\frac72\right)
\quad\text{and}\quad
\left(\frac32,\frac72,\frac{11}{2},\frac{11}{2},\frac72,\frac32\right),
\]
whose Pearson correlation, and hence Spearman correlation of the underlying RDMs, is $1/2$. Therefore
\[
\mathcal{S}(Y)
=
\frac{1}{3}\left(
\frac12-\frac12+\frac12
\right)
=
\frac16
\neq
\frac12
=
\mathcal{S}(X).
\]

Because $Q$ is orthogonal,
\[
YY^\top
=
XQQ^\top X^\top
=
XX^\top.
\]
Hence the rowwise inner products and norms are also identical, so the full cosine-distance RDM is the same for $X$ and $Y$. Thus linear $\operatorname{CKA}(X,Y)=1$ exactly, and an RSA comparison based on these full cosine-distance RDMs likewise cannot distinguish the two representations, while $\mathcal{S}(Y)\neq\mathcal{S}(X)$. Therefore Shesha\textsubscript{FS} is not invariant under general orthogonal transformations of the feature basis.
\end{proof}

\paragraph{Scope.}
This counterexample establishes that an orthogonal transformation can alter $\mathcal{S}$ while preserving $XX^\top$ exactly. It does not imply that every orthogonal transformation changes $\mathcal{S}$ or that orthogonal rotations must change it in a particular direction.

\subsubsection{Numerical Confirmation at Fixed Spectrum}
The constructive example above is deliberately small and hand-verifiable. The same dissociation occurs in a higher-dimensional stable representation. We generated $X \in \mathbb{R}^{200 \times 64}$ by projecting a five-dimensional latent ($Z \in \mathbb{R}^{200 \times 5}$, standard normal) through a dense random map ($W \in \mathbb{R}^{5 \times 64}$) with small additive noise (seed 320), so that geometric structure is redundantly distributed across coordinates and the representation has high feature-split stability ($\widehat{\mathcal{S}}_{30}(X)=0.903$). Rotating $X$ into its right-singular-vector basis, $Y=XQ$, is an orthogonal transformation and therefore leaves $XX^\top$, the singular spectrum, the rank, and linear CKA unchanged ($\operatorname{CKA}(X,Y)=1.000$). Yet $\widehat{\mathcal{S}}_{30}(Y)=-0.008$. Rotating into the right-singular-vector basis concentrates the dominant variance associated with the five-dimensional latent into a small number of individual coordinates. Random coordinate halves therefore receive different subsets of these dominant directions, so the two half-RDMs can reflect different latent structure even though the full Gram matrix is unchanged. This numerical example illustrates the basis dependence established exactly by the counterexample above.

\subsection{Strictly Monotone RDM-Transform Invariance}
\label{app:proof-monotone}
\begin{proof}
Spearman correlation is the Pearson correlation of rank vectors. Let $g:\mathbb{R}\to\mathbb{R}$ be strictly monotone and apply the same $g$ elementwise to both half-RDMs.

If $g$ is strictly increasing, it preserves the ordering, including tie classes, of the RDM entries. Thus
\[
\operatorname{rank}(g(D))=\operatorname{rank}(D)
\]
and the split-half Spearman correlation is unchanged.

If $g$ is strictly decreasing, it reverses the ordering. Writing $m=n(n-1)/2$ for the number of upper-triangular RDM entries, average ranks obey
\[
\operatorname{rank}(g(D))
=
(m+1)\mathbf{1}-\operatorname{rank}(D).
\]
Applying this affine reversal to both rank vectors leaves their Pearson correlation unchanged:
\[
\operatorname{corr}\!\left(
(m+1)\mathbf{1}-r_A,\,
(m+1)\mathbf{1}-r_B
\right)
=
\operatorname{corr}(r_A,r_B).
\]
Therefore every split statistic, and hence both $\mathcal{S}$ and a fixed-partition $\widehat{\mathcal{S}}_K$, are invariant to applying the same strictly monotone transformation to the RDM distance values.
\end{proof}

\section{Connection to the RSA Noise Ceiling}
\label{si:noise_ceiling}

Shesha is related conceptually to the RSA noise ceiling of \citet{Nili2014}, but the two procedures assess different forms of reliability. In the standard RSA noise-ceiling construction, each subject's RDM is computed from the full multivariate representation. The upper-bound estimate compares each subject RDM with the group-average RDM including that subject, whereas the lower-bound estimate compares it with the average RDM of the remaining subjects. The resulting ceiling quantifies how well a model could be expected to predict the empirical representational geometry given measurement variability across subjects.

Shesha also measures agreement between RDMs, but it changes the representation from which each RDM is constructed. For a single representation $X \in \mathbb{R}^{n\times d}$, Shesha partitions the $d$ feature coordinates into complementary subsets $A$ and $B$ and computes separate RDMs from $X_A$ and $X_B$. Each RDM therefore observes only part of the feature population. Shesha asks whether these incomplete coordinate subsets recover similar pairwise sample geometry.

The distinction can be summarized as follows:
\begin{enumerate}
    \item \textit{Source of variation}
    \begin{itemize}
        \item Noise ceiling: variation across repeated estimates of the
        representation, typically across subjects.
        \item Shesha: variation induced by which feature coordinates of a
        single representation are retained.
    \end{itemize}

    \item \textit{Representation used to construct each RDM}
    \begin{itemize}
        \item Noise ceiling: each RDM is constructed from the full available
        multivariate feature population for that replicate.
        \item Shesha: each RDM is constructed from a complementary subset of
        the feature population.
    \end{itemize}

    \item \textit{Diagnostic target}
    \begin{itemize}
        \item Noise ceiling: reliability of the estimated representational
        geometry under measurement variability.
        \item Shesha: consistency of representational geometry across subsets
        of the feature basis.
    \end{itemize}

    \item \textit{Data requirements}
    \begin{itemize}
        \item Noise ceiling: requires multiple estimates of the
        representational geometry, such as multiple subjects.
        \item Shesha: requires only a single representation matrix
        $X \in \mathbb{R}^{n\times d}$.
    \end{itemize}
\end{enumerate}

Thus the connection is methodological rather than an identity of procedures. Both quantify agreement between representational dissimilarity structures, but the noise ceiling asks whether the \emph{full} representational geometry is reproducible across measurements, whereas Shesha asks whether that geometry is recoverable when the feature population itself is divided. Statistical properties or interpretations of the RSA noise ceiling therefore do not automatically transfer to Shesha.

\section{Shesha Computation}
\label{si:shesha_computation}

All Shesha\textsubscript{FS} computations followed a standardized protocol. Feature dimensions were randomly partitioned into two disjoint halves of equal size (for odd $d$, one half received $(d+1)/2$ features). Cosine distance RDMs were computed from each half using Eq.~\ref{eq:rdm}. Spearman rank correlation between the vectorized upper triangles of the two RDMs was then computed. This procedure was repeated for $K{=}30$ independent random partitions and the results averaged.

When $n^2$ RDM computation was prohibitive, samples were subsampled to $n_{\max} = 1{,}600$ (stratified by available labels when present, random otherwise). Convergence analysis across 15 models on CIFAR-10 and CIFAR-100 found a mean absolute difference of 0.0115 between estimates at $n = 400$ and $n = 1{,}600$ across the 30 model--dataset combinations (Sec.~\ref{si:convergence}). We use $n_{\max} = 1{,}600$ throughout as a conservative ceiling.

All computations used fixed random seed 320 for reproducibility, unless otherwise noted. Float64 precision was used throughout for ranking and correlation computations to reduce numerical ties introduced by finite precision.

CKA was computed as debiased linear CKA using the unbiased HSIC estimator~\citep{song2012feature}. This estimator removes diagonal contributions and applies the corresponding finite-sample correction, reducing the positive finite-sample bias that can arise for independent high-dimensional representations under standard linear CKA~\citep{kornblith2019similarity}.

\section{Synthetic and Controlled Validation}
\label{si:ground-truth}

This section examines Shesha\textsubscript{FS} on synthetic and controlled data where the data-generating parameters are known. The aim is narrower than treating Shesha as an alternative similarity measure: we test whether it responds systematically to controlled changes in feature-split structure and whether it can vary independently of between-representation similarity.

\begin{figure}[H]
\centering
\includegraphics[width=.5\textwidth]{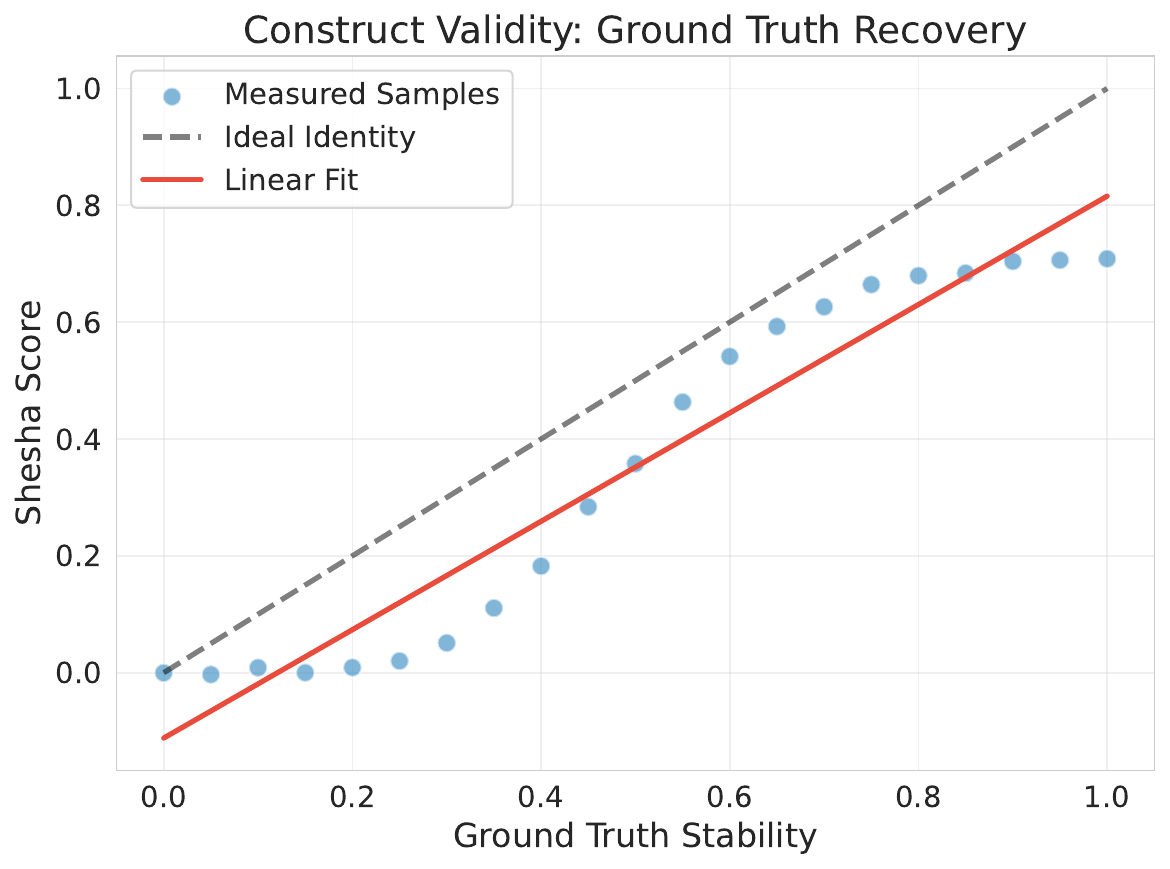}
\caption{Synthetic signal-to-noise sweep: Shesha\textsubscript{FS} scores plotted against the mixing parameter $\alpha$ in controlled representations. The metric shows a near-perfect monotonic response ($\rho = 0.997$) across the sweep.}
\label{fig:S1}
\end{figure}

We first use a signal-to-noise sweep (Sec.~\ref{si:sensitivity}, Fig.~\ref{fig:S1}), where Shesha\textsubscript{FS} is almost perfectly monotonic with the mixing parameter ($\rho = 0.997$). We then use a balanced four-quadrant construction (Sec.~\ref{app:balanced-quad}, Fig.~\ref{fig:S2}) in which high and low Shesha values are paired with both high and low CKA values. Across this balanced sample, the Spearman correlation between Shesha\textsubscript{FS} and debiased CKA is $\rho = 0.204$, showing that Shesha is not a monotonic re-expression of CKA in this construction. The encoder-transformation analysis of the main text (Sec.~\ref{sec:validation}) provides a complementary check in the sense of \citet{kornblith2019similarity} and \citet{ding2021grounding}: some transformations move the two statistics together, whereas PCA-coordinate compression produces a regime in which their responses differ. Because Shesha\textsubscript{FS} is basis-dependent (Appendix~\ref{si:proofs}), such differences are expected when a transformation changes how geometry is distributed across coordinates.

The remaining subsections assess numerical convergence, seed sensitivity, and responses to controlled spectral manipulations.

\subsection{Sample-Size Convergence}
\label{si:convergence}
To assess whether Shesha estimates converge reliably as sample size
varies, we measured $\mathcal{S}$ at two sample sizes
($n \in \{400, 1600\}$) across 15 models on both CIFAR-10 and
CIFAR-100~\citep{Krizhevsky09learningmultiple}. For each
model-dataset combination, we randomly sampled $n$ examples without
replacement using a fixed random generator and measured the drift
$\Delta = \mathcal{S}_{n=400} - \mathcal{S}_{n=1600}$. Stability
was defined as $|\Delta| < 0.05$.

\begin{figure}[H]
\centering
\includegraphics[width=.6\textwidth]{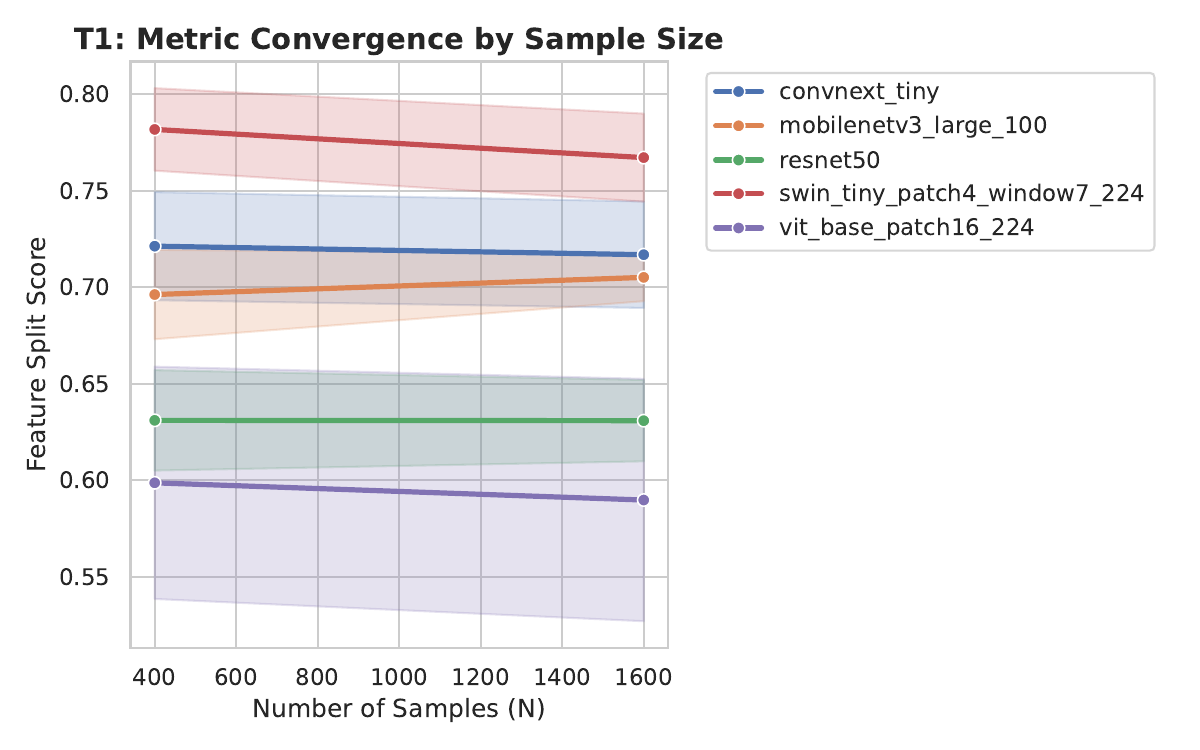}
\caption{Sample-size convergence: Shesha estimates at $n=400$ and $n=1{,}600$ across representative architectures. All model--dataset combinations satisfy the prespecified $|\Delta|<0.05$ criterion.}
\label{fig:convergence}
\end{figure}

Shesha estimates demonstrated excellent convergence across all
architectures (Fig.~\ref{fig:convergence}). The mean absolute drift
across all 30 model-dataset combinations was
$|\bar{\Delta}| = 0.0115$, well below the stability threshold. When
averaged per model across both datasets, drifts ranged from 0.0002
(ResNet-50, most stable) to 0.0176 (ViT-Tiny, least stable), with
mean 0.0077. All 15 models satisfied the prespecified $|\Delta|<0.05$ criterion at $n = 400$. We use $n_{\max} = 1{,}600$ throughout as a
conservative ceiling and $K = 30$ splits as the estimation protocol.

\subsection{Dimensionality Sensitivity}
To assess how Shesha\textsubscript{FS} behaves under dimensionality
reduction, which is common in visualization and computational-efficiency
contexts, we applied Principal Component Analysis (PCA) to reduce embeddings
from their native dimensionality (512--2048, depending on the architecture)
to 64 dimensions. For each of the 30 model-dataset combinations, we
extracted 400 samples, fit PCA with \texttt{n\_components=64} and
\texttt{random\_state=320}, transformed the embeddings, and recomputed
Shesha\textsubscript{FS} on the reduced representations.

Reduction to 64 components drove Shesha\textsubscript{FS} from a full-dimensional mean of $+0.620$ to a negative mean of $-0.112$ (range $[-0.204, -0.055]$ across the 30 conditions). Thus, in these representations, PCA-coordinate compression substantially changes feature-split stability. The formal result in Appendix~\ref{si:proofs-pca} is deliberately narrower: Shesha is not invariant to PCA-coordinate projection, and a projection can change Shesha even when no Gram information is lost. It does not imply that every PCA projection must reduce Shesha. Because the present study is concerned with stability in the learned feature basis, all main analyses use the native, full-dimensional representations unless a dimensionality manipulation is itself the object of study.

\subsection{Sensitivity to Known Stability Levels}
\label{si:sensitivity}

We generated representations with parametrically controlled stability by mixing a low-rank signal component with isotropic noise:
\begin{equation*}
    X = \alpha \cdot \frac{Z W}{\|ZW\|_F} + (1 - \alpha) \cdot \epsilon
\end{equation*}
where $Z \in \mathbb{R}^{n \times k}$ is a latent matrix ($n=200$ samples, $k=50$ latent dimensions), $W \in \mathbb{R}^{k \times d}$ is a random projection ($d=256$ features), $\epsilon \sim \mathcal{N}(0, I)$ is isotropic noise, and $\alpha \in [0, 1]$ controls the mixture between shared low-rank signal and isotropic noise. We tested 21 levels from $\alpha = 0$ to $\alpha = 1$ in increments of 0.05, using seeds $\mathcal{S}[i \mod 15] \times 100 + i$ for each level $i \in \{0, \ldots, 20\}$. Shesha showed near-perfect rank correlation with this controlled signal-to-noise parameter ($\rho = 0.997$).

\subsection{Spectral Deletion}
\label{si:spectral-deletion}

To compare how the statistics respond to spectral deletion, we constructed representations with a power-law eigenspectrum ($S_{ii} = 100/(i+1)$) and progressively removed the top $k$ principal components. The similarity measures compare each modified representation with the original representation, whereas Shesha\textsubscript{FS} measures feature-split agreement within the modified representation itself.

After removal of the first principal component, CKA, PWCKA, and Procrustes all fall below 0.5, while Shesha declines more gradually and remains positive at $k = 50$ (Table~\ref{tab:spectral}). This experiment therefore demonstrates different responses to the same spectral manipulation; it does not treat the statistics as estimates of the same underlying quantity.

\begin{table}[H]
\centering
\caption{Metric values after removing the top $k$ principal components. Similarity metrics compare each modified representation with the original; Shesha reports feature-split agreement within the modified representation. $k$ = PCs removed. {$^a$Shesha at $k=0$ reflects feature-split agreement rather than self-similarity.}}
\label{tab:spectral}
\begin{tabular}{lcccccc}
\hline
$k$ & Shesha & CKA & Debiased CKA & PWCKA & Procrustes \\
\hline

0  & 0.981$^a$ & 1.000 & 1.000 & 1.000 & 1.000 \\
1  & 0.955 & 0.273 & 0.262 & 0.274 & 0.389 \\
2  & 0.932 & 0.136 & 0.118 & 0.136 & 0.238 \\
3  & 0.905 & 0.083 & 0.060 & 0.083 & 0.170 \\
4  & 0.876 & 0.057 & 0.031 & 0.057 & 0.132 \\
5  & 0.850 & 0.043 & 0.012 & 0.043 & 0.108 \\
6  & 0.823 & 0.033 & 0.000 & 0.033 & 0.091 \\
7  & 0.802 & 0.027 & $-$0.009 & 0.027 & 0.078 \\
8  & 0.774 & 0.022 & $-$0.016 & 0.022 & 0.069 \\
9  & 0.744 & 0.019 & $-$0.022 & 0.019 & 0.061 \\
10 & 0.717 & 0.016 & $-$0.027 & 0.016 & 0.055 \\
\vdots\\
15 & 0.607 & 0.009 & $-$0.045 & 0.009 & 0.036 \\
20 & 0.495 & 0.006 & $-$0.058 & 0.006 & 0.027 \\
25 & 0.392 & 0.004 & $-$0.067 & 0.004 & 0.021 \\
30 & 0.309 & 0.003 & $-$0.075 & 0.002 & 0.017 \\
35 & 0.238 & 0.003 & $-$0.082 & 0.002 & 0.014 \\
40 & 0.171 & 0.002 & $-$0.087 & 0.001 & 0.012 \\
45 & 0.124 & 0.002 & $-$0.092 & 0.000 & 0.010 \\
50 & 0.086 & 0.001 & $-$0.097 & 0.000 & 0.009 \\
\hline
\end{tabular}
\end{table}
\normalsize

\subsection{Tail Noise Ablation}
\label{si:tail_noise}

The spectral-deletion experiment (Table~\ref{tab:spectral}) shows that Shesha\textsubscript{FS} and similarity-to-original measures respond differently when leading principal components are removed. As a complementary control, we asked how the same statistics respond when Gaussian noise is injected only into components outside the subspace explaining 90\% of the variance.

\subsubsection{Protocol}
We generated synthetic representations
$X \in \mathbb{R}^{2000 \times 512}$ with a power-law eigenspectrum
($\lambda_i \propto i^{-1.5}$, matching the spectral profile of trained
deep network penultimate layers; seed 320). We decomposed $X$ via SVD
and identified the top $k = 34$ principal components explaining 90\% of
total variance as the signal subspace. We then injected isotropic
Gaussian noise exclusively into the remaining 478 tail components at
14 scale levels ($\sigma \in [0.001, 50]$), reconstructed the
perturbed representation $\tilde{X}$ in the original coordinate space,
and measured three diagnostics: linear CKA between $X$ and $\tilde{X}$, RDM similarity between $X$ and $\tilde{X}$ (Spearman correlation
of pairwise distance vectors), and internal Shesha\textsubscript{FS} of $\tilde{X}$.
Each condition was repeated with 3 independent noise draws.

\subsubsection{Results}
At low noise scales ($\sigma \leq 0.01$), CKA and RDM similarity remain above 0.999, and internal Shesha\textsubscript{FS} remains at its baseline value of 0.971. As noise increases, the cross-comparison metrics degrade: CKA drops below 0.95 at $\sigma = 0.20$, while RDM similarity crosses the same threshold at $\sigma = 0.10$. At $\sigma = 0.05$, both remain above 0.99. At high noise scales ($\sigma \geq 1.0$), where the injected tail energy is large relative to the original tail variance, both cross-comparison metrics approach zero.

Across this sweep, Shesha does not show a separate low-noise response that is absent from CKA. Thus the response pattern seen under leading-component deletion is not reproduced by this tail-noise manipulation.

\subsubsection{Interpretation}
The two manipulations answer different control questions. Deleting leading components changes the represented subspace directly; tail-noise injection perturbs the retained representation without removing those components. The invariance result in Sec.~\ref{sec:invariances} applies to monotone transformations of RDM entries themselves and should not be read as an invariance to additive feature noise. Here, the empirical point is simply that modest tail-noise injection leaves Shesha nearly unchanged, while sufficiently large perturbations alter it along with the cross-representation similarity measures.

\subsection{Preprocessing Ablation}
\label{si:preprocessing}
Following \cite{Walther2016}, we repeated the spectral-deletion experiment under four preprocessing conditions: raw, centered, centered with L2 normalization, and whitened (ZCA with shrinkage parameter $0.1$). In every condition, debiased CKA against the original representation is near zero or negative after removing the top $k=30$ principal components, while Shesha\textsubscript{FS} remains positive (Table~\ref{tab:preproc}). Shesha\textsubscript{FS} changes with centering, as expected for a statistic built on cosine distance, which is not translation-invariant. Whitening lowers the Shesha\textsubscript{FS} baseline from $0.98$ to $0.50$ at $k=0$, consistent with noise amplification when variance is equalized across directions, but leaves the qualitative pattern at $k=30$ unchanged.

\begin{table}[H]
\centering
\caption{Shesha\textsubscript{FS} and debiased CKA after removing the top
$k=30$ principal components under different preprocessing. CKA compares each
modified representation with the original; Shesha\textsubscript{FS} is
computed within the modified representation.}
\label{tab:preproc}
\begin{tabular}{lcc}
\hline
Preprocessing & Shesha\textsubscript{FS} & Debiased CKA \\
\hline
Raw                   & 0.276 & $-$0.076 \\
Centered              & 0.417 & $-$0.076 \\
Centered + Normalized & 0.417 & $-$0.083 \\
Whitened              & 0.316 & $-$0.054 \\
\hline
\end{tabular}
\end{table}

\subsubsection{Mechanistic Interpretation of Whitening} Under whitening, CKA remains negative at $k=30$ ($-0.054$), though less so than under raw preprocessing ($-0.076$). The whitened Shesha baseline drops from 0.98 to 0.50 at $k=0$, reflecting noise amplification from spectral equalization.

\subsection{Comparison with RSA Reliability Methods} We additionally compared standard Shesha with a whitened-RDM variant motivated by RSA reliability methods~\citep{Walther2016, diedrichsen2017representational} and considered the conceptual relation to noise-ceiling procedures~\citep{Nili2014}. Standard and whitened Shesha are almost perfectly rank-correlated in this analysis ($\rho=1.000$, $p<10^{-70}$), indicating that whitening does not materially change the ordering of conditions here. Unlike whitening-based procedures, the definition used in the main analyses operates directly on the reported representation coordinates and does not require ZCA preprocessing.

Taken together, these controls show that the spectral-deletion dissociation is robust to several preprocessing choices. The formal distinction is basis-dependence: Appendix~\ref{si:proofs} gives an orthogonal transformation that preserves the full Gram matrix while changing Shesha\textsubscript{FS}.

\begin{figure}[H]
\centering
\includegraphics[width=\textwidth]{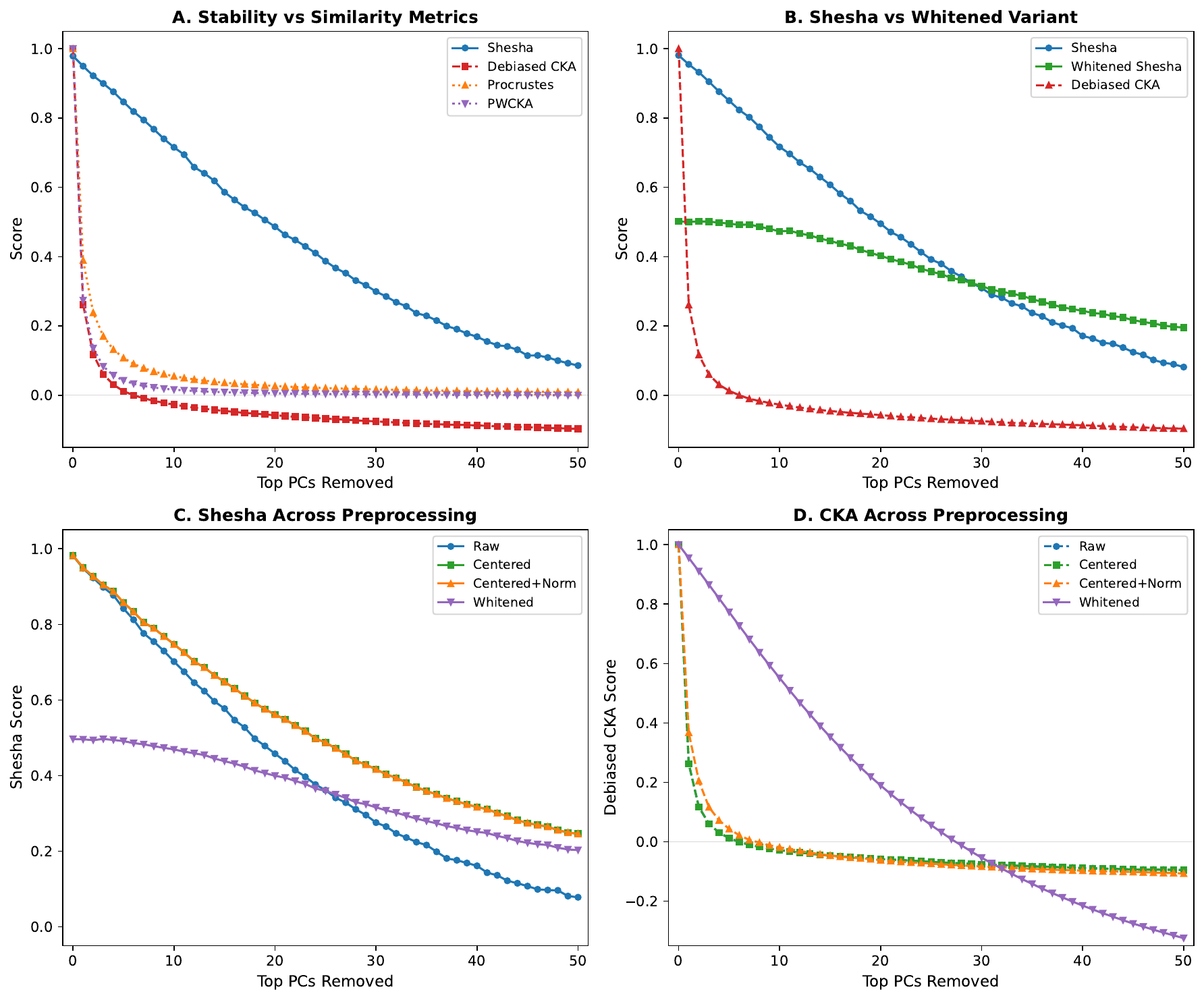}
\caption{Spectral and preprocessing controls. {(A)} Responses as the top $k$ principal components are progressively removed from a power-law representation. Similarity-to-original measures fall sharply after removal of the leading component, while Shesha declines more gradually. {(B)} Comparison with whitened Shesha ($\rho = 0.999$). {(C)} Shesha across preprocessing conditions (raw, centered, normalized, whitened). {(D)} CKA across the same preprocessing conditions.}
\label{fig:spectral}
\end{figure}

\subsection{Seed Stability}

To verify that the stochastic feature-splitting procedure produces consistent estimates across random initializations, we computed $\mathcal{S}$ twice for each of 15 models on both CIFAR-10 and CIFAR-100, using \texttt{seed=100} and \texttt{seed=200} respectively. Each seed generates a different sequence of $K = 30$ random feature partitions. Sensitivity was measured as $|\mathcal{S}_{\text{seed}=100} - \mathcal{S}_{\text{seed}=200}|$.

\begin{figure}[H]
\centering
\includegraphics[width=.45\textwidth]{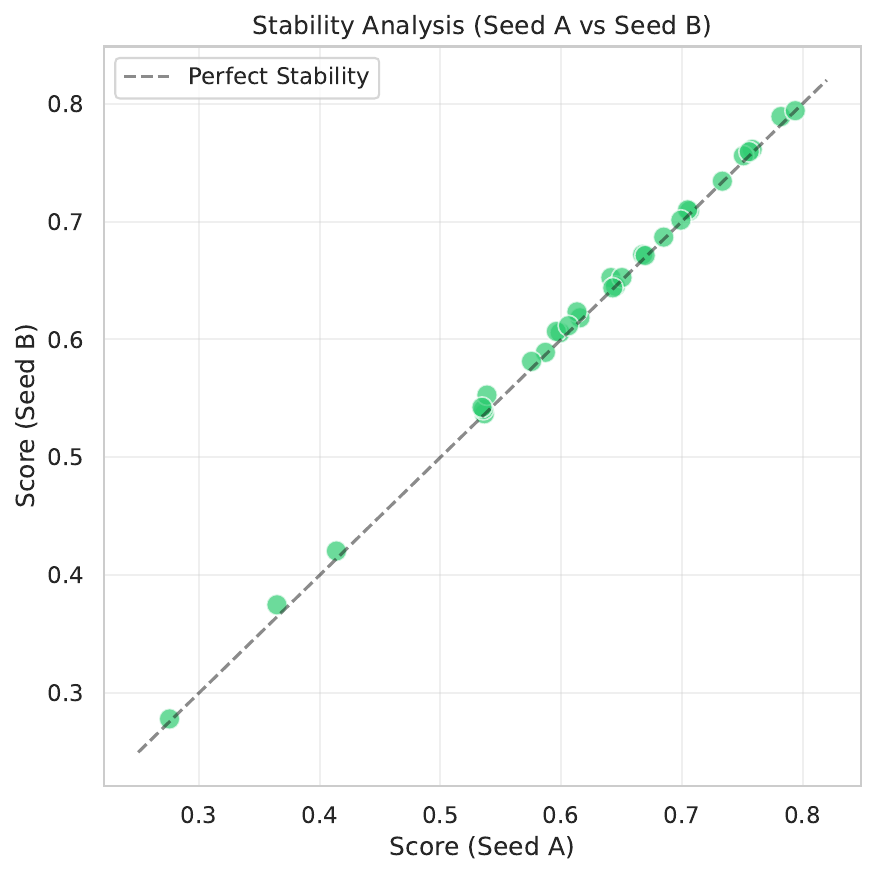}
\caption{Seed stability: Shesha scores computed with
\texttt{seed=100} vs.\ \texttt{seed=200} across 15 architectures on
CIFAR-10 and CIFAR-100. Points align closely with the diagonal,
confirming high reproducibility across random initializations. Mean
sensitivity $= 0.0047$; maximum $= 0.0142$.}
\label{fig:stability}
\end{figure}

The metric demonstrated excellent seed stability across all architectures and datasets (Fig.~\ref{fig:stability}). The mean sensitivity across all 30 model-dataset combinations was 0.0047, with a maximum of 0.0142 (ResNet-34 on CIFAR-100) and a minimum of 0.00015 (ResNet-50 on CIFAR-10). All 30 combinations fell well below the 0.05 stability threshold, with 25/30 below 0.01. These results confirm that averaging over $K = 30$ random splits provides sufficient variance reduction to yield highly reproducible estimates, with typical seed-to-seed variation below 1\% of the score magnitude.

\subsection{Dissociation with Balanced Quadrant Sampling}
\label{app:balanced-quad}

\begin{figure}[H]
\centering
\includegraphics[width=0.7\textwidth]{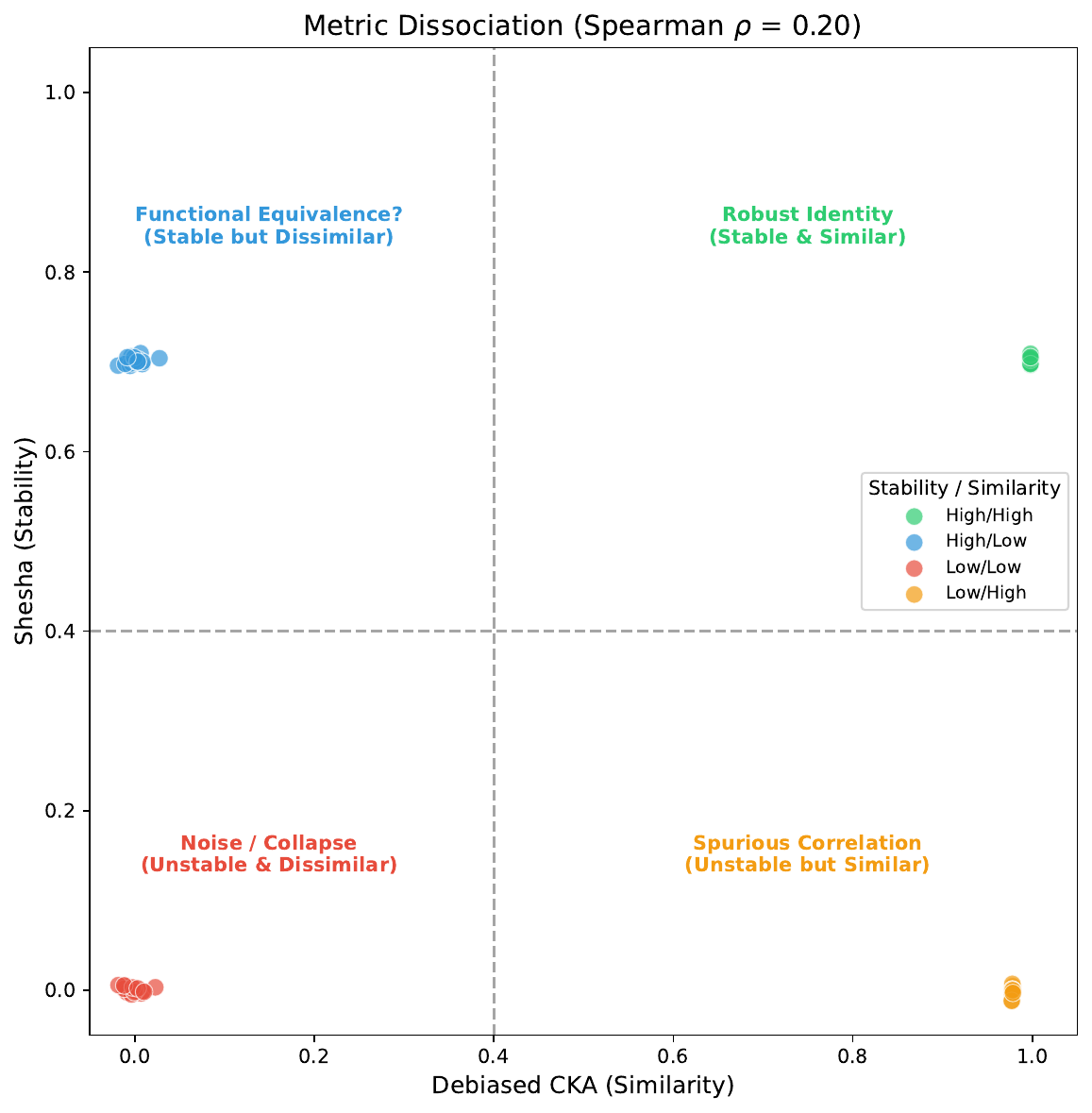}
\caption{Four-quadrant dissociation: Shesha vs.\ debiased
CKA for 60 representation pairs sampled equally from four quadrants
of the stability $\times$ similarity space. Q1 (high/high): Shesha
$= 0.701 \pm 0.003$, CKA $= 0.998 \pm 0.000$. Q2 (high/low): Shesha
$= 0.701 \pm 0.004$, CKA $= 0.001 \pm 0.010$. Q3 (low/low): Shesha
$= 0.001 \pm 0.003$, CKA $= -0.001 \pm 0.010$. Q4 (low/high,
adversarial): Shesha $= -0.001 \pm 0.005$, CKA $= 0.978 \pm 0.000$.
Balanced Spearman $\rho = 0.20$.}
\label{fig:S2}
\end{figure}

In this synthetic construction, unconstrained random sampling can couple Shesha and CKA because the signal strength and the relationship between the paired latent draws both affect the observed statistics. To sample the two quantities more evenly, we explicitly constructed four quadrants (15 pairs each, using seeds derived from $\mathcal{S}$):
\begin{enumerate}
    \item \textbf{High stability, high similarity} (Q1): Representations derived from the same latent structure ($\alpha=0.9$) with small additive noise ($\sigma=0.1$). Seeds: $\mathcal{S}[i] \times 1000 + 1$ for $i \in \{1, \ldots, 15\}$. Results: Shesha $= 0.701 \pm 0.003$, CKA $= 0.998 \pm 0.000$.
    
    \item \textbf{High stability, low similarity} (Q2): Independent high-signal representations ($\alpha=0.9$) with different latent draws. Seeds: $\mathcal{S}[i] \times 1000 + 2$ and $\mathcal{S}[i] \times 1000 + 3$ for each pair. Results: Shesha $= 0.701 \pm 0.004$, CKA $= 0.001 \pm 0.010$.
    
    \item \textbf{Low stability, low similarity} (Q3): Independent noise representations ($\alpha=0.1$). Seeds: $\mathcal{S}[i] \times 1000 + 4$ and $\mathcal{S}[i] \times 1000 + 5$ for each pair. Results: Shesha $= 0.001 \pm 0.003$, CKA $= -0.001 \pm 0.010$.
    
    \item \textbf{Low stability, high similarity} (Q4): Adversarial quadrant constructed via rejection sampling. We generated pairs where $X \sim \mathcal{N}(0, I)^{200 \times 256}$ and $Y = X + \mathcal{N}(0, 0.15^2 I)$, accepting only samples where Shesha $< 0.4$ and CKA $> 0.4$. This creates representations with aligned sample geometry (high CKA) but inconsistent feature-split structure (low Shesha). Acceptance rate: 100\% (15/15). Results: Shesha $= -0.001 \pm 0.005$, CKA $= 0.978 \pm 0.000$.
\end{enumerate}
With equal numbers of samples from each quadrant, the Spearman correlation between Shesha and debiased CKA is $\rho = 0.204$ (Fig.~\ref{fig:S2}). The construction shows directly that high or low feature-split stability can coexist with either high or low between-representation similarity; neither statistic is a monotonic function of the other in this design.

\section{Encoders: Extended Methods and Results}
\label{si:encoders}
The stability--similarity analysis in Sec.~\ref{sec:validation} is evaluated on a heterogeneous corpus rather than a single model or dataset. This appendix specifies the 2{,}463 configurations spanning seven data domains, including data sources, encoders, and preprocessing.

\subsection{Cross-Domain Validation: Data Sources and Preprocessing}
\label{si:domains}
To assess whether the observed relationship depends strongly on modality, the corpus spans seven domains ranging from natural language to neural population recordings. Within each domain we fix a stimulus set and enumerate configurations by varying the encoder or encoding scheme and its preprocessing; the resulting counts $N$ and full protocols are given below.

\subsubsection{Language (\texorpdfstring{$N{=}127$}{N=127})}
Sentences from the SST-2 validation set~\citep{socher-etal-2013-recursive} were tokenized using each model's default tokenizer with padding and truncation (max length: 64 tokens). Representations were extracted from the final hidden layer and mean-pooled across tokens using attention masks. 500 sentences; base models: \texttt{all-MiniLM-L6-v2}, \texttt{all-mpnet-base-v2}, \texttt{distilbert-base-nli-stsb-mean tokens}, and \texttt{paraphrase-distilroberta-base-v1}.

\subsubsection{Vision (\texorpdfstring{$N{=}129$}{N=129})}
Images from CIFAR-100~\citep{Krizhevsky09learningmultiple} were preprocessed using each model's default image processor (resized to $224 \times 224$, ImageNet normalization). Representations were extracted from the final layer with global average pooling. 400 images; base models: \texttt{google/vit-base-patch16-224}, \texttt{openai/clip-vit-base-patch32}, \\\texttt{facebook/deit-base-patch16-224}, and ResNet50 (ImageNet-V2 weights).

\subsubsection{Audio (\texorpdfstring{$N{=}64$}{N=64})}
Audio samples from LibriSpeech dev-clean~\citep{panayotov2015librispeech} were resampled to 16\,kHz and truncated/padded to 1 second duration. Representations were extracted from the final encoder layer and mean-pooled across time. 200 samples; base models: \\\texttt{facebook/wav2vec2-base-960h} and \texttt{facebook/hubert-base-ls960}.

\subsubsection{Video (\texorpdfstring{$N{=}128$}{N=128})}
Action clips were drawn from UCF-101~\citep{Soomro2012UCF101AD}: 100 videos were sampled uniformly at random (seed-controlled) from the on-disk corpus, with 16 frames per clip selected by uniform temporal indexing and resized to $224 \times 224$ with ImageNet normalization. Base models comprised two temporal transformers\\(\texttt{facebook/timesformer-base-finetuned-k400} - 8 frames, \texttt{MCG-NJU/videomae-base} - 16 frames) and two frame-level encoders: ViT-B/16 (\texttt{google/vit-base-patch16-224}) applied to the temporal mean frame, and CLIP ViT-B/32 (\texttt{openai/clip-vit-base-patch32}) with embeddings from four uniformly spaced frames averaged per clip. 

A preliminary analysis using 100 clips uniformly sampled from a single-source Jellyfish video~\citep{allyn2016jellyfish} with the same base models and preprocessing gave a similar correlation ($\rho = -0.24$ vs.\ $\rho = -0.27$), providing an additional video-source check.

\subsubsection{Protein (\texorpdfstring{$N{=}402$}{402})}
Protein sequences from Swiss-Prot (UniProt reviewed human proteins; \citealt{uniprot2023uniprot}), filtered to lengths between 50 and 2,000 residues. 200 sequences; multiple encoding schemes: amino acid composition (20-dim), dipeptide frequency (400-dim), hydrophobicity and charge profiles at multiple resolutions (25, 50, 100 bins), and 3-mer spectra (500-dim hashed).

\subsubsection{Molecular (\texorpdfstring{$N{=}767$}{N=767})}
Single-cell RNA-seq data from the pbmc3k dataset~\citep{Zheng2017}, loaded with Scanpy\\~\citep{Wolf2018}. Genes with fewer than 3 expressing cells were filtered. 1,000 cells; multiple preprocessing strategies: log-transformation, various PCA dimensions, top-variance gene selection, CPM normalization, and binarization (presence/absence).

\subsubsection{Neural Population Recordings (\texorpdfstring{$N{=}846$}{N=846})}
Neuropixels recordings from~\citet{steinmetz2019distributed}, comprising high-density recordings from 29,134 neurons across 68 brain areas in awake mice. Sessions were filtered to include only those with at least 20 neurons and 50 trials ($N{=}26$ qualified sessions). Spike counts were binned at 20\,ms resolution and averaged across time bins.

\subsection{Encoder Transformations}
\label{si:transformations}

For each base representation in each domain, we applied a standardized set of geometric interventions, resulting in 2,463 unique encoder configurations across all seven domains, aggregated across 15 seeds ($3, 7, 9, 11, 12, 18, 103, 108, 320, 411, 724, 1754, 1991, 2222,$ $7258$). 

\begin{figure}[h!]
\centering
\resizebox{\textwidth}{!}{%
\begin{tikzpicture}[
    font=\sffamily,
    base_rep/.style={rectangle, draw=black!80, fill=gray!10, thick,
        minimum width=3cm, minimum height=1.5cm, rounded corners=3pt, align=center},
    transform_box/.style={rectangle, minimum width=3cm, minimum height=1cm,
        align=center, font=\footnotesize},
    mechanism_box/.style={rectangle, draw=black!40, thick, minimum width=3.2cm,
        minimum height=1.0cm, rounded corners=2pt, align=center, font=\footnotesize, fill=white},
    outcome_box/.style={rectangle, draw=black!60, thick, minimum width=3.8cm,
        minimum height=1.2cm, rounded corners=2pt, align=left, font=\footnotesize},
    arrow/.style={->, >=stealth, very thick}
]

\node[base_rep] (input) at (0,0) {Base Representation\\ $X \in \mathbb{R}^{N \times D}$};

\node[transform_box, text=blue!70!black] (trans_jl) at (4.5, 1.6)
    {Random Projection\\ (Johnson-Lindenstrauss)};
\node[transform_box, text=violet] (trans_pca) at (4.5, -1.6)
    {PCA Compression\\ (variance concentration)};

\node[mechanism_box] (mech_jl) at (8.9, 1.6)
    {Approximately preserves\\ pairwise distances};
\node[mechanism_box] (mech_pca) at (8.9, -1.6)
    {Changes dimension and\\ coordinate basis};

\node[outcome_box, fill=blue!8] (outcome_jl) at (13.6, 1.6)
    {CKA high, Shesha high\\ responses track};
\node[outcome_box, fill=violet!10] (outcome_pca) at (13.6, -1.6)
    {CKA and Shesha\\ responses diverge};

\draw[arrow, color=blue!60!black] (input.east) -- (2.2, 0) -- (2.2, 1.6) -- (trans_jl.west);
\draw[arrow, color=violet] (input.east) -- (2.2, 0) -- (2.2, -1.6) -- (trans_pca.west);
\draw[arrow, color=blue!60!black] (trans_jl.east) -- (mech_jl.west);
\draw[arrow, color=violet] (trans_pca.east) -- (mech_pca.west);
\draw[arrow, dashed, color=blue!60!black] (mech_jl.east) -- (outcome_jl.west);
\draw[arrow, dashed, color=violet] (mech_pca.east) -- (outcome_pca.west);

\end{tikzpicture}
}
\caption{Two response regimes in the encoder-transformation analysis. Under random projection, CKA and Shesha tend to remain high together. Under PCA-coordinate compression, their responses diverge. The pooled near-zero correlation therefore reflects a mixture of transformation regimes rather than uniformly weak within-regime associations.}
\label{fig:regimes}
\end{figure}

\subsubsection{PCA}
Principal component projection to $k$ dimensions, with $k \in \{5, 10, \ldots, 300\}$ (capped at $\min(n, d) - 1$).

\subsubsection{Random Projection}
Gaussian random projection to $k$ dimensions, $k \in \{16, 32, \ldots, 256\}$.

\subsubsection{Top-Variance Feature Selection}
Selection of $k$ features with highest marginal variance, $k \in \{50, 100, \ldots, 800\}$.

\subsubsection{Random Feature Subsets}
Random subset of $k$ features without replacement, $k \in \{50, 100, 200\}$.

\subsubsection{Gaussian Noise Injection}
Additive Gaussian noise scaled by $\sigma \cdot \mathrm{std}(X)$, with $\sigma \in \{0.05, 0.1, \ldots, 1.0\}$.

\subsubsection{Normalization}
Z-score (per-feature zero mean, unit variance) and L2 (per-sample unit norm).

\subsection{Similarity Metrics}
\label{si:similarity}

For each encoder configuration, CKA was computed between the transformed representation and three domain-specific reference representations: the original untransformed base representation, a PCA projection at $k{=}100$ (or the closest available rank), and a z-scored version. The three CKA values were averaged to produce a single similarity score per configuration, minimizing single-reference artifacts.

Alternative similarity metrics were evaluated in the language domain ($N{=}127$) and are reported in Table~\ref{tab:S7}.

\subsubsection{Effective-Rank Projection-Weighted CKA (PWCKA)}
This variant projects both representations to a shared dimensionality determined by the effective rank before computing CKA. Given the centered representations $\mathbf{X}, \mathbf{Y} \in \mathbb{R}^{n \times d}$, we compute their singular value decompositions:
\begin{equation*}
    \mathbf{X} = \mathbf{U}_X \mathbf{S}_X \mathbf{V}_X^\top, \quad \mathbf{Y} = \mathbf{U}_Y \mathbf{S}_Y \mathbf{V}_Y^\top
\end{equation*}
The effective rank $k$ is the minimum number of components explaining 99\% of variance in either representation:
\begin{equation*}
    k = \min\left( k_X^{(0.99)}, k_Y^{(0.99)} \right), \quad \text{where } k_Z^{(\tau)} = \min \left\{ j : \frac{\sum_{i=1}^{j} s_z^{(i)2}}{\sum_{i} s_z^{(i)2}} \geq \tau \right\}
\end{equation*}
CKA is then computed on the truncated projections:
\begin{equation*}
    \mathbf{X}' = \mathbf{U}_X^{(1:k)} \mathbf{S}_X^{(1:k)}, \quad \mathbf{Y}' = \mathbf{U}_Y^{(1:k)} \mathbf{S}_Y^{(1:k)}
\end{equation*}
\begin{equation*}
    \text{PWCKA}(\mathbf{X}, \mathbf{Y}) = \text{CKA}(\mathbf{X}', \mathbf{Y}')
\end{equation*}

\subsubsection{Procrustes Similarity}
Procrustes analysis finds the optimal orthogonal transformation that aligns two representations. Given centered representations $\mathbf{X}, \mathbf{Y} \in \mathbb{R}^{n \times d}$, we first normalize them to a unit Frobenius norm:
\begin{equation*}
    \tilde{\mathbf{X}} = \frac{\mathbf{X}}{\|\mathbf{X}\|_F}, \quad \tilde{\mathbf{Y}} = \frac{\mathbf{Y}}{\|\mathbf{Y}\|_F}
\end{equation*}
The optimal orthogonal matrix $\mathbf{R}^* = \underset{\mathbf{R}^\top\mathbf{R} = \mathbf{I}}{\arg\min} \|\tilde{\mathbf{X}} - \tilde{\mathbf{Y}}\mathbf{R}\|_F^2$ is obtained via the SVD of the cross-covariance matrix:
\begin{equation*}
    \tilde{\mathbf{Y}}^\top \tilde{\mathbf{X}} = \mathbf{U}\mathbf{\Sigma}\mathbf{V}^\top \implies \mathbf{R}^* = \mathbf{U}\mathbf{V}^\top
\end{equation*}
Procrustes similarity is defined as follows:
\begin{equation*}
    \text{Procrustes}(\mathbf{X}, \mathbf{Y}) = 1 - \frac{\|\tilde{\mathbf{X}} - \tilde{\mathbf{Y}}\mathbf{R}^*\|_F^2}{\|\tilde{\mathbf{X}}\|_F^2 + \|\tilde{\mathbf{Y}}\mathbf{R}^*\|_F^2}
\end{equation*}

\begin{table}[H]
\caption{Alternative similarity metrics in the language domain. All three show $|\rho| < 0.30$ with Shesha in this analysis.}
\label{tab:S7}
\centering
\begin{tabular}{lcc}
\hline
Similarity metric & $\rho$ with Shesha & $p$ \\
\hline
CKA & $-0.01$ & 0.74 \\
PWCKA          & $-0.22$ & 0.012 \\
Procrustes     & $+0.28$ & 0.001 \\
\hline
\end{tabular}
\end{table}

\subsection{Statistical Methods}
\label{si:statistics}
This subsection specifies the inferential procedures behind the distinctness analysis (Sec.~\ref{sec:validation}) and the vision benchmark. We describe the resampling scheme used for confidence intervals, the mixed-effects control for base-model identity, the group comparisons, and our handling of multiple comparisons.

\subsubsection{Bootstrap Inference}
Distinctness was assessed via Spearman rank correlation with 10,000 bootstrap replicates, resampling encoder configurations within each domain. 95\% confidence intervals were computed as bootstrap percentile intervals.

\subsubsection{Mixed-Effects Models}
To assess sensitivity to base-model identity, we fit a mixed-effects model with $\mathcal{S}$ as outcome, debiased CKA as a fixed effect, and base model as a random intercept. The intraclass correlation coefficient (ICC) for base model was 0.10, indicating that about 10\% of the variance in stability was associated with between-model differences in this model. The fixed-effect slope of CKA on $\mathcal{S}$ was $-0.03$ (95\% CI $[-0.08, +0.02]$), consistent with the aggregate near-zero association reported in the main text.

\subsubsection{Mann-Whitney U Tests}
Architectural comparisons (contrastive vs.\ self-supervised; hierarchical vs.\ columnar) used two-sided Mann-Whitney U tests on Shesha\textsubscript{FS} scores, reported with exact $p$-values.

\subsubsection{Multiple Comparisons}
Per-dataset statistical tests in the vision benchmark are reported without multiplicity correction and should be interpreted as dataset-specific analyses rather than as familywise-error-controlled evidence across all datasets.

\begin{figure}[H]
\centering
\includegraphics[width=0.45\textwidth]{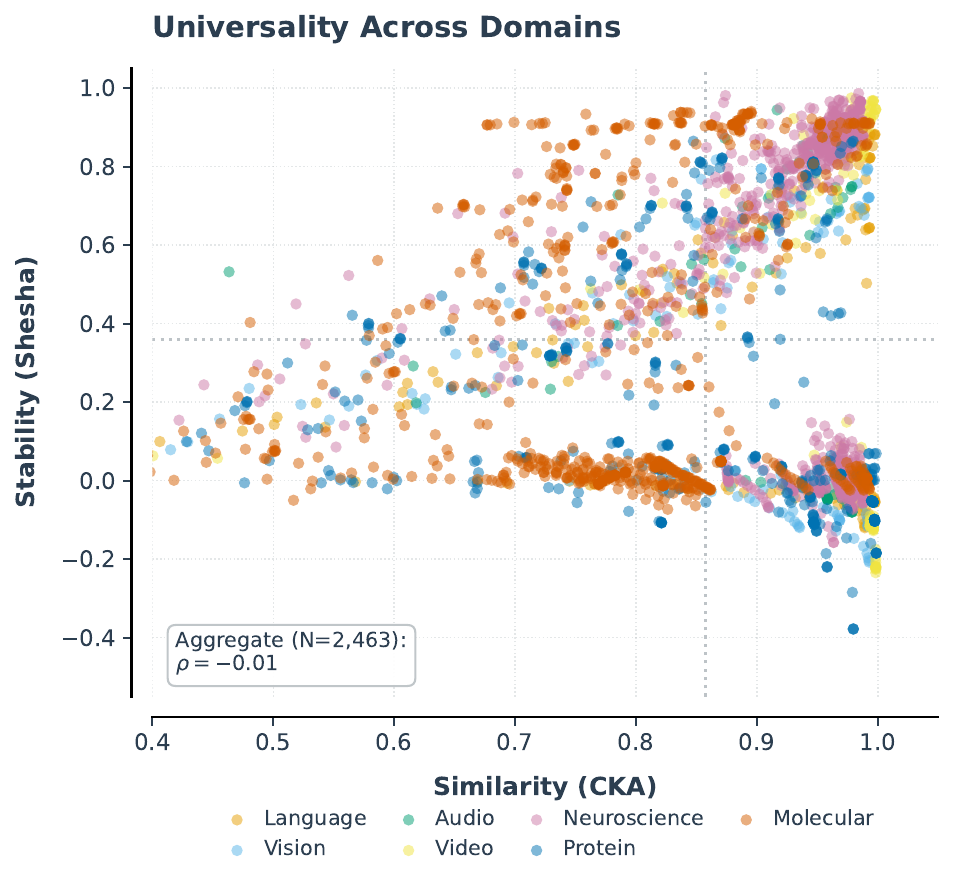}
\caption{Across 2,463 encoder configurations spanning seven domains, Shesha and CKA show negligible pooled rank correlation ($\rho = -0.01$, 95\% CI $[-0.06, +0.03]$).}
\label{fig:universal}
\end{figure}

\begin{table}[H]
\centering
\caption{Correlation between Shesha and CKA by encoder type.}
\label{si:regime-analysis}
\begin{tabular}{@{}lcc@{}}
\hline
{Encoder Type} & \textit{N} & $\rho$ [95\% CI] \\
\hline
Random Features & 201 & $+0.92$ [$+0.89$, $+0.94$] \\
Random Projection & 395 & $+0.89$ [$+0.86$, $+0.92$] \\
Noise Injection & 395 & $+0.58$ [$+0.49$, $+0.66$] \\
Top Variance & 287 & $+0.62$ [$+0.54$, $+0.71$] \\
Normalization & 158 & $+0.34$ [$+0.17$, $+0.50$] \\
\hline
Original & 79 & $+0.31$ [$+0.04$, $+0.55$] \\
\hline
PCA & 948 & $-0.47$ [$-0.52$, $-0.42$] \\
\hline
\end{tabular}
\end{table}

\begin{table}[H]
\caption{Robustness checks for aggregate distinctness.
All subsets maintain $|\rho| < 0.10$.}
\label{tab:S5}
\centering
\begin{tabular}{lrc}
\hline
Analysis & $N$ & $\rho$ [95\% CI] \\
\hline
Full dataset & 2463 & $-0.01$ [$-0.06$, $+0.03$] \\
Excluding Neuroscience & 1617 & $-0.09$ [$-0.14$, $-0.04$] \\
Excluding Protein & 2061 & $+0.04$ [$-0.00$, $+0.09$] \\
Only transformer domains & 448 & $-0.05$ [$-0.15$, $+0.07$] \\
Only biological domains & 2015 & $+0.01$ [$-0.04$, $+0.06$] \\
\hline
\end{tabular}
\end{table}

\begin{table}[h!]
\caption{Domain-level correlations between stability and similarity. Aggregate correlation is negligible ($\rho = -0.01$, CI within $\pm 0.06$); four domains show negligible correlations ($|\rho| < 0.10$). $^{a}$Protein shows moderate negative correlation driven by PCA on low-dimensional sequence encoders
(20--500 dims).% ; see Appendices~\ref{si:domains}--\ref{si:transformations}.
}
\label{tab:domains}
\centering
\small
\begin{tabular}{lrcrc}
\hline
Domain & $N$ & $\rho$ & 95\% CI & $p$ \\
\hline
\multicolumn{5}{l}{\textit{Machine Learning}} \\
\quad Language     & 127  & $+0.03$ & $[-0.18, +0.24]$ & 0.77 \\
\quad Vision       & 129  & $-0.03$ & $[-0.23, +0.18]$ & 0.72 \\
\quad Audio        &  64  & $-0.26$ & $[-0.52, +0.02]$ & 0.04 \\
\quad Video        & 128  & $-0.27$ & $[-0.47, -0.05]$ & 0.002 \\[4pt]
\multicolumn{5}{l}{\textit{Biology}} \\
\quad Neuroscience & 846  & $+0.01$ & $[-0.06, +0.09]$ & 0.67 \\
\quad Protein$^{a}$& 402  & $-0.36$ & $[-0.45, -0.28]$ & ${<}0.001$ \\
\quad Molecular    & 767  & $+0.06$ & $[-0.02, +0.13]$ & 0.13 \\[4pt]
\hline
{Aggregate} & {2463} & ${-0.01}$ &
    ${[-0.06, +0.03]}$ & {0.57} \\
\hline
\end{tabular}
\end{table}

\section{SAM vs.\ SGD Ablation: Full Protocol and Extended Results}
\label{si:sam_ablation}
Sharpness-Aware Minimization (SAM;~\citealt{foret2021sharpnessaware}) optimizes against parameter perturbations within an $\ell_2$-ball of radius $\rho$ before each update. We use the SAM radius as a controlled training intervention and ask whether Shesha\textsubscript{FS} and CKA respond in the same way as $\rho$ changes.

\subsection{Protocol}
We trained ResNet-18 models on CIFAR-10 and CIFAR-100 using identical hyperparameters: SGD base optimizer with learning rate 0.05, momentum 0.9, weight decay $5 \times 10^{-4}$, batch size 128, cosine annealing over 100
epochs. The only variable was the SAM perturbation radius $\rho \in \{0, 0.01, 0.02, 0.05, 0.1, 0.2\}$, where $\rho = 0$ recovers standard SGD. Each configuration was trained over 15 random seeds. The ResNet-18 architecture was adapted for CIFAR with a $3 \times 3$ initial convolution (stride 1, padding 1) and identity max-pooling. After training, we extracted 512-dimensional penultimate-layer representations (post-average-pooling) for 2{,}000 test images, and computed test accuracy, debiased linear CKA against the SGD baseline, Shesha\textsubscript{FS} ($K = 30$ splits), and three supervised Shesha variants (variance ratio, supervised alignment, class separation ratio), which we introduced in another work~\citep{raju2026canary}. The label-aware variants are: the variance ratio (between-class to total variance), a supervised alignment score, and a class-separation ratio; all three increase with the concentration of variance along class-discriminative directions and serve here only as a contrast to the unsupervised Shesha\textsubscript{FS}. All metrics are reported as mean\,$\pm$\,SD over the 15 seeds ($3, 7, 9, 11, 12, 18, 103, 108, 320, 411, 724, 1754, 1991, 2222,$ $7258$).

\subsection{Results}
Tables~\ref{tab:sam_ablation_c10} and~\ref{tab:sam_ablation_c100} report all metrics. Three patterns hold.

First, the intervention dissociates Shesha\textsubscript{FS} from CKA. As $\rho$ increases, CKA against the SGD baseline falls steadily, from 1.000 to 0.925 on CIFAR-10 and, more sharply, from 1.000 to 0.772 by $\rho = 0.01$ on CIFAR-100, where it then stays near 0.77. Shesha\textsubscript{FS} does not follow this decline. At the peak radius it exceeds the SGD baseline on every one of the 15 seeds on both datasets (CIFAR-10, $\rho = 0.05$: $+0.067$, paired $t_{14} = 25.3$, $p = 4 \times 10^{-13}$; CIFAR-100, $\rho = 0.2$: $+0.018$, $t_{14} = 16.6$, $p = 1 \times 10^{-10}$; two-sided). Thus, across a single controlled training sweep, accuracy changes modestly while CKA and Shesha\textsubscript{FS} show different response profiles.

Second, the observed maximum is dataset-dependent rather than monotone across the sweep. On CIFAR-10 Shesha\textsubscript{FS} peaks at $\rho = 0.05$--$0.1$ (0.872) and declines at $\rho = 0.2$ (0.851). The variance ratio changes in the same part of the sweep, reaching its minimum at $\rho = 0.1$ (0.786) and rising slightly at $\rho = 0.2$ (0.790). The decline at the largest CIFAR-10 radius is compatible with non-monotonic behavior reported for SAM at larger perturbation radii~\citep{andriushchenko2022towards}, although the present experiment does not identify its cause. On CIFAR-100 the increase is shallower and continues through the largest tested radius, reaching 0.822 at $\rho = 0.2$. We report both shapes rather than averaging across datasets.

Third, the label-aware quantities do not track Shesha\textsubscript{FS}. On CIFAR-10 the variance ratio declines from 0.849 to 0.790 and the class-separation ratio from 3.08 to 2.41 across the sweep, while supervised alignment stays approximately flat ($\approx 0.51$); CIFAR-100 shows the same qualitative pattern. The controlled sweep therefore separates feature-split stability from these supervised summary statistics without requiring either quantity to serve as a proxy for the other.

\begin{table}[H]
\centering
\caption{SAM ablation, CIFAR-10 (full results), mean\,$\pm$\,SD over 15 seeds.}
\label{tab:sam_ablation_c10}
\resizebox{\textwidth}{!}{%
\begin{tabular}{lcccccc}
\hline
$\rho$ & Test Acc.\,(\%) & CKA vs.\ SGD & Shesha\textsubscript{FS}
& Var.\ Ratio & Sup.\ Align. & Class Sep. \\
\hline
0.00 (SGD) & $94.91 \pm 0.15$ & 1.000 & $0.806 \pm 0.008$ & $0.849 \pm 0.004$ & $0.511 \pm 0.004$ & $3.08 \pm 0.05$ \\
0.01       & $95.07 \pm 0.13$ & $0.949 \pm 0.001$ & $0.831 \pm 0.008$ & $0.830 \pm 0.004$ & $0.511 \pm 0.004$ & $2.81 \pm 0.04$ \\
0.02       & $95.29 \pm 0.19$ & $0.946 \pm 0.001$ & $0.851 \pm 0.005$ & $0.817 \pm 0.003$ & $0.511 \pm 0.004$ & $2.65 \pm 0.03$ \\
0.05       & $95.46 \pm 0.08$ & $0.938 \pm 0.002$ & $0.872 \pm 0.007$ & $0.796 \pm 0.004$ & $0.511 \pm 0.004$ & $2.46 \pm 0.03$ \\
0.10       & $95.56 \pm 0.14$ & $0.933 \pm 0.002$ & $0.872 \pm 0.008$ & $0.786 \pm 0.003$ & $0.511 \pm 0.003$ & $2.38 \pm 0.02$ \\
0.20       & $95.50 \pm 0.12$ & $0.925 \pm 0.003$ & $0.851 \pm 0.020$ & $0.790 \pm 0.003$ & $0.511 \pm 0.004$ & $2.41 \pm 0.02$ \\
\hline
\end{tabular}
}
\end{table}

\begin{table}[H]
\centering
\caption{SAM ablation, CIFAR-100 (full results), mean\,$\pm$\,SD over 15 seeds.}
\label{tab:sam_ablation_c100}
\resizebox{\textwidth}{!}{%
\begin{tabular}{lcccccc}
\hline
$\rho$ & Test Acc.\,(\%) & CKA vs.\ SGD & Shesha\textsubscript{FS}
& Var.\ Ratio & Sup.\ Align. & Class Sep. \\
\hline
0.00 (SGD) & $76.96 \pm 0.21$ & 1.000 & $0.805 \pm 0.003$ & $0.562 \pm 0.003$ & $0.159 \pm 0.004$ & $1.505 \pm 0.005$ \\
0.01       & $77.11 \pm 0.25$ & $0.772 \pm 0.003$ & $0.805 \pm 0.003$ & $0.542 \pm 0.003$ & $0.158 \pm 0.005$ & $1.468 \pm 0.006$ \\
0.02       & $77.32 \pm 0.22$ & $0.773 \pm 0.002$ & $0.805 \pm 0.003$ & $0.533 \pm 0.002$ & $0.158 \pm 0.004$ & $1.452 \pm 0.004$ \\
0.05       & $77.48 \pm 0.20$ & $0.777 \pm 0.002$ & $0.806 \pm 0.003$ & $0.520 \pm 0.003$ & $0.158 \pm 0.004$ & $1.430 \pm 0.005$ \\
0.10       & $77.82 \pm 0.24$ & $0.775 \pm 0.002$ & $0.814 \pm 0.004$ & $0.504 \pm 0.002$ & $0.157 \pm 0.005$ & $1.406 \pm 0.004$ \\
0.20       & $78.05 \pm 0.20$ & $0.765 \pm 0.002$ & $0.822 \pm 0.003$ & $0.485 \pm 0.002$ & $0.156 \pm 0.005$ & $1.381 \pm 0.004$ \\
\hline
\end{tabular}
}
\end{table}

\section{Vision Benchmark: Extended Methods and Results}
\label{si:vision-extended}

This appendix specifies the vision benchmark behind Sec.~\ref{sec:vision}: how the 170 models were selected, and how their features and transferability scores were computed. The subsections that follow document these choices and the extended analyses that support the main-text claims.

\subsection{Model Selection}
170 pretrained vision models were drawn from the PyTorch Image Models (timm) library\\\citep{rw2019timm}. Selection ensured broad coverage across four axes: (i) training objectives (supervised ImageNet-1k/21k, self-supervised DINO/DINOv2/MAE, contrastive CLIP, generative EVA-02/BEiT); (ii) architectural families (columnar ViT/DeiT, hierarchical Swin/SwinV2/PVT-v2, hybrid CoAtNet/MaxViT, convolutional ResNet/ConvNeXt/\\EfficientNet/RegNet/DenseNet); (iii) model scales (MobileNetV3-Small to ViT-Giant/14); (iv) training paradigms (standard, distillation, augmentation, foundation model pretraining). Models were grouped into 36 semantic families for aggregate analysis. When training objective and architecture conflicted, training objective was prioritized for family assignment (e.g., ViT-CLIP assigned to ``CLIP'' rather than ``ViT'').

\subsection{Feature Extraction}
Penultimate-layer features were extracted from fixed random subsets of each dataset (seed 320): 5,000 images for CIFAR-10, CIFAR-100, and EuroSAT; 5,000 for Flowers-102 (with replacement where the dataset is smaller); 1,500 for Oxford Pets; 1,600 for DTD.
All images were preprocessed using each model's standard transform (resize, center crop, normalization).

\subsection{Transferability Metrics}
LogME \citep{you_ranking_2022, you2021logme} was computed on the same features using the authors' implementation. LEEP \citep{nguyen2020leep} was computed for models with classification heads.

\subsection{Extended Results}

\begin{figure}[H]
\centering
\includegraphics[width=\textwidth]{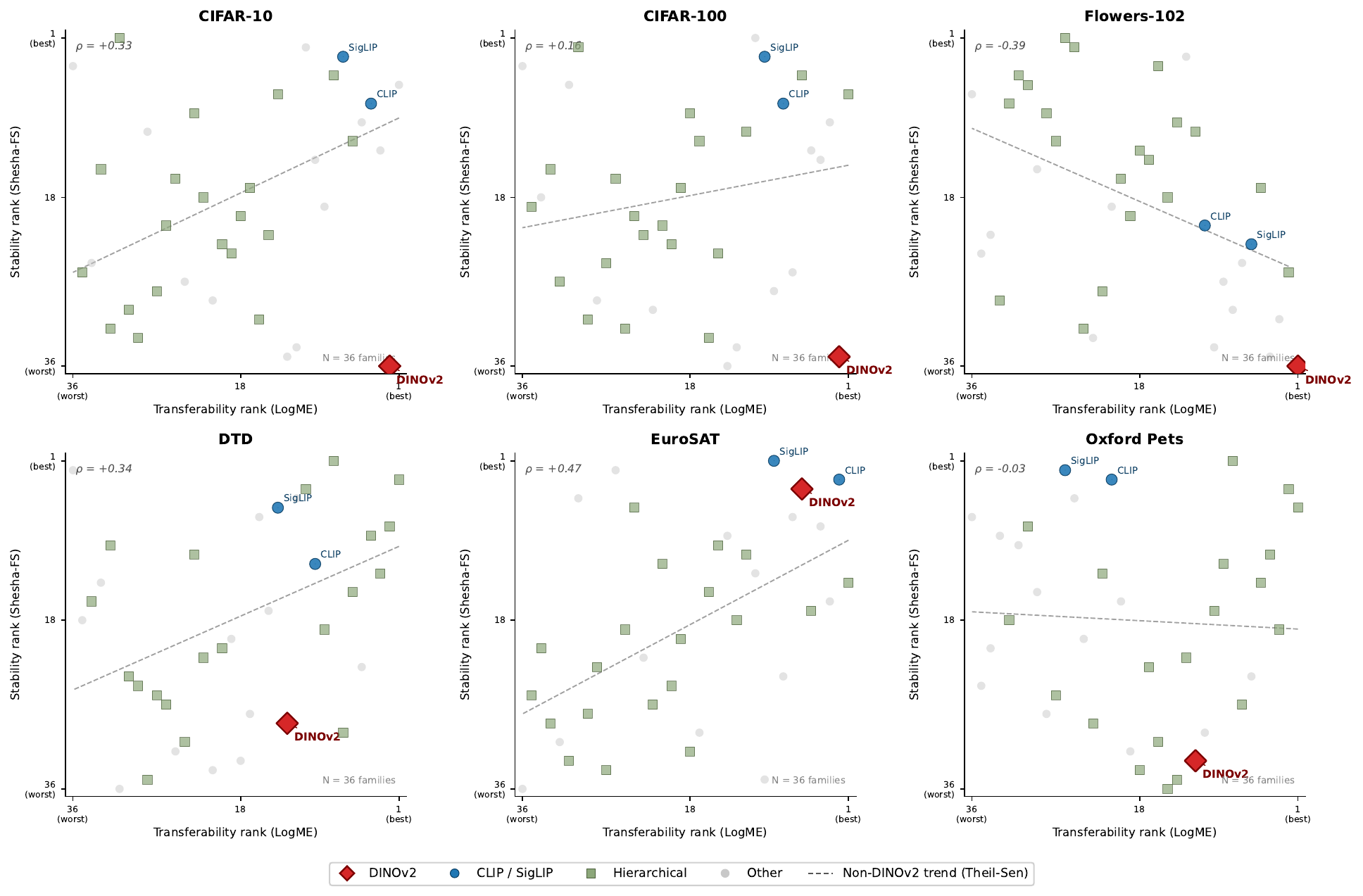}
\caption{Per-dataset family rankings underlying Fig.~\ref{fig:dino}A. Each panel plots family-mean transferability rank (LogME) against geometric-stability rank (Shesha\textsubscript{FS}) for the 36 families, with both axes oriented worst to best. DINOv2 occupies the high-transfer, low-stability region on five datasets, but not on EuroSAT, where it ranks among the most stable families. The robust Theil-Sen cross-family trend is positive or near zero on five datasets (CIFAR-10 $+0.33$, CIFAR-100 $+0.16$, DTD $+0.34$, EuroSAT $+0.47$, Oxford Pets $-0.03$) and negative on Flowers-102 ($-0.39$). The relationship between transferability and Shesha therefore varies across datasets rather than showing a consistent trade-off.}
\label{fig:family_grid}
\end{figure}

\begin{table}[H]
\caption{DINOv2 at the individual-model level. DINOv2-giant ranks in
the bottom quartile for Shesha\textsubscript{FS} on every dataset except EuroSAT,
while attaining top-six transferability on CIFAR-10, CIFAR-100, and Flowers-102.
On EuroSAT it is instead among the most stable models (4/170).}
\label{tab:S2}
\centering
\begin{tabular}{lcccc}
\hline
Dataset & LogME & LogME Rank & Shesha\textsubscript{FS} & $\mathcal{S}$ Rank \\
\hline
CIFAR-10    & 1.384 & 6/170  & 0.414 & 160/170 \\
CIFAR-100   & 1.629 & 3/170  & 0.319 & 158/170 \\
Flowers-102 & 3.521 & 3/170  & 0.152 & 168/170 \\
DTD         & 0.952 & 30/170 & 0.502 & 129/170 \\
EuroSAT     & 0.681 & 16/170 & 0.987 & 4/170 \\
Oxford Pets & 1.760 & 30/170 & 0.569 & 141/170 \\
\hline
\end{tabular}
\end{table}

\begin{table}[H]
\caption{Contrastive vs.\ self-supervised stability: Mann-Whitney U tests
comparing contrastive models (CLIP, SigLIP, ViTamin; $n{=}29$) to self-supervised
models ($n{=}41$) on Shesha\textsubscript{FS}. $^*p < 0.05$; $^{**}p < 0.01$;
$^{***}p < 0.001$. Contrastive models have higher Shesha\textsubscript{FS} scores on all six datasets.}
\label{tab:S3}
\centering
\begin{tabular}{lcccc}
\hline
Dataset & Contrastive ($n{=}29$) & SSL ($n{=}41$) & $\Delta$ & $p$ \\
\hline
CIFAR-10    & $0.80 \pm 0.05$ & $0.67 \pm 0.17$ & $+0.13$ & ${<}0.001^{***}$ \\
CIFAR-100   & $0.76 \pm 0.08$ & $0.61 \pm 0.21$ & $+0.15$ & ${<}0.001^{***}$ \\
Flowers-102 & $0.83 \pm 0.03$ & $0.69 \pm 0.24$ & $+0.14$ & $0.012^{*}$ \\
DTD         & $0.68 \pm 0.08$ & $0.58 \pm 0.14$ & $+0.09$ & ${<}0.001^{***}$ \\
EuroSAT     & $0.95 \pm 0.01$ & $0.89 \pm 0.07$ & $+0.06$ & ${<}0.001^{***}$ \\
Oxford Pets & $0.86 \pm 0.03$ & $0.69 \pm 0.12$ & $+0.17$ & ${<}0.001^{***}$ \\
\hline
\end{tabular}
\end{table}

\begin{table}[H]
\caption{Hierarchical vs.\ columnar stability: Mann-Whitney U tests on
Shesha\textsubscript{FS} comparing hierarchical, multi-scale architectures
(e.g.\ Swin, PVT, CoAtNet, and convolutional networks; $n{=}81$) to columnar,
single-scale ones (e.g.\ ViT, DeiT; $n{=}89$). $^*p < 0.05$; $^{**}p < 0.01$;
$^{***}p < 0.001$. The advantage is significant only on Flowers-102.}
\label{tab:S4}
\centering
\begin{tabular}{lcccc}
\hline
Dataset & Hier.\ ($n{=}81$) & Col.\ ($n{=}89$) & $\Delta$ & $p$ \\
\hline
CIFAR-10    & $0.71 \pm 0.09$ & $0.70 \pm 0.16$ & $+0.01$ & $0.891$ \\
CIFAR-100   & $0.67 \pm 0.10$ & $0.64 \pm 0.20$ & $+0.03$ & $0.646$ \\
Flowers-102 & $0.86 \pm 0.11$ & $0.75 \pm 0.19$ & $+0.11$ & ${<}0.001^{***}$ \\
DTD         & $0.61 \pm 0.14$ & $0.61 \pm 0.14$ & $+0.00$ & $0.582$ \\
EuroSAT     & $0.86 \pm 0.06$ & $0.90 \pm 0.11$ & $-0.04$ & $1.000$ \\
Oxford Pets & $0.67 \pm 0.13$ & $0.74 \pm 0.14$ & $-0.07$ & $1.000$ \\
\hline
\end{tabular}
\end{table}

\subsection{Seed Stability of Vision Models}
\label{si:seed}

The Shesha\textsubscript{FS} estimator averages over $K = 30$ random feature partitions, and the partition sequence is fixed by a random seed. To assess sensitivity to this Monte Carlo choice, we recomputed the entire CIFAR-10 sweep under three independent seeds (9, 320, and 1991), each generating a different set of $K = 30$ partitions, for all 170 models. For each model we summarized the three per-seed scores by their mean, standard
deviation, and coefficient of variation (CV, standard deviation divided by mean).

Rankings are nearly invariant to the seed. Between any pair of seeds, Shesha\textsubscript{FS} rankings agree at Kendall $\tau \in [0.944, 0.945]$ and raw values at Spearman $\rho \in [0.993, 0.995]$; LogME is even more stable ($\tau \in [0.988, 0.991]$, $\rho \geq 0.9996$). Per-model variation is small: the median Shesha\textsubscript{FS} CV is 0.75\% (LogME 1.08\%), and only 5 of 170 models (2.9\%) exceed a 5\% CV (3 of 170, 1.8\%, for LogME). The five high-variability models are mid-to-low-stability architectures (deit3\_base, vit\_tiny, vit\_base.mae, wide\_resnet50\_2, densenet201) whose small absolute fluctuations inflate the ratio; in every case the absolute Shesha\textsubscript{FS} standard deviation across seeds is below 0.06.

The low-Shesha ranking of DINOv2 is also stable across partition seeds. The register-augmented large DINOv2 variant holds rank 170 of 170 under all three seeds (Shesha\textsubscript{FS} $= 0.291, 0.287, 0.297$), and the DINOv2 family records the lowest family-mean Shesha\textsubscript{FS} under every seed.

A Friedman test across the 170 paired models does detect a small systematic difference between seeds for Shesha\textsubscript{FS} ($\chi^2 = 8.48$, $p = 0.014$; for LogME $\chi^2 = 175.78$, $p < 10^{-6}$). This reflects the sensitivity of a paired test at $n = 170$ to sub-percent shifts rather than any practical instability: the effect is negligible in magnitude (median CV 0.75\%) and leaves the rankings essentially unchanged. We therefore report seed 320 throughout the main text and treat the metric as reproducible across partition seeds.

\begin{table}[h]
\caption{CIFAR-10 seed stability across three feature-partition seeds
(9, 320, 1991) for all 170 models. Median Shesha\textsubscript{FS} and LogME per
seed, and Spearman $\rho$ between seeds on raw values. A Friedman test detects a
small systematic seed effect for Shesha\textsubscript{FS} ($\chi^2 = 8.48$,
$p = 0.014$); its magnitude is negligible (median per-model CV 0.75\%) and rank
agreement is near-perfect ($\rho \geq 0.993$, Kendall $\tau \geq 0.944$).}
\label{tab:seed_stability}
\centering
\begin{tabular}{l|c|c|c}
\hline
Metric & Seed 9 & Seed 320 & Seed 1991 \\
\hline
Median LogME & 0.5486 & 0.5615 & 0.5566 \\
Median Shesha\textsubscript{FS} & 0.7167 & 0.7137 & 0.7161 \\
\hline
Spearman $\rho$ (LogME) & 9 vs 320: 0.9997 & 9 vs 1991: 0.9996 & 320 vs 1991: 0.9996 \\
Spearman $\rho$ (Shesha\textsubscript{FS}) & 9 vs 320: 0.9944 & 9 vs 1991: 0.9930 & 320 vs 1991: 0.9948 \\
\hline
\end{tabular}
\end{table}

\subsection{Probe Subset-Sensitivity: Full Protocol and Robustness Analysis}
\label{si:probe_sensitivity}

This subsection tests a narrower practical prediction of feature-split stability. If complementary feature subsets induce less similar representational geometries, analyses that depend on different subsets of the feature coordinates may also be less consistent. We test this using linear probes across 170 vision models and six datasets. Because downstream probe variability also depends strongly on task alignment, we treat LogME as the primary control and report mean-accuracy and coefficient-of-variation analyses as secondary checks.

\subsubsection{Protocol}

For each of 170 vision models, we evaluated clean representations on CIFAR-10, CIFAR-100, Flowers-102, DTD, EuroSAT, and Oxford Pets. Within each dataset, the train/test split was held fixed across all probes so that variation in test accuracy reflected the choice of feature subset rather than the choice of examples. For each model, we repeatedly selected a random half of the feature dimensions, standardized those features using the training split, trained a logistic-regression probe, and recorded test accuracy.

The primary analysis used $K=100$ random feature halves per model and dataset. We also repeated the full analysis with $K=20$ feature halves as a robustness check. For each model, we summarized probe performance by the mean, standard deviation, range, and coefficient of variation of test accuracy across feature subsets. Probe subset-sensitivity is quantified primarily by the standard deviation, $\sigma_{\mathrm{probe}}$. Shesha\textsubscript{FS} was computed on the corresponding full clean representation.

\subsubsection{Cross-Dataset Results}

The marginal relationship between Shesha\textsubscript{FS} and probe subset-sensitivity is heterogeneous across datasets. At $K=100$, Shesha\textsubscript{FS} is negatively associated with $\sigma_{\mathrm{probe}}$ on CIFAR-10 ($\rho=-0.294$, $p=9.77\times10^{-5}$), CIFAR-100 ($\rho=-0.337$, $p=7.02\times10^{-6}$), DTD ($\rho=-0.405$, $p=4.24\times10^{-8}$), and EuroSAT ($\rho=-0.690$, $p=2.38\times10^{-25}$). The marginal association is weak on Oxford Pets ($\rho=-0.117$, $p=0.127$) and positive on Flowers-102 ($\rho=+0.379$, $p=3.56\times10^{-7}$). Thus, Shesha\textsubscript{FS} alone does not determine downstream probe subset-sensitivity. 

Because probe variability is also strongly related to task alignment, our primary conditional analysis controls for LogME. At $K=100$, the partial Spearman correlation \\$\rho(\mathrm{Shesha},\sigma_{\mathrm{probe}}\mid\mathrm{LogME})$ is negative on five of six datasets: CIFAR-10 ($-0.318$, $p=2.49\times10^{-5}$), CIFAR-100 ($-0.513$, $p=1.03\times10^{-12}$), DTD ($-0.371$, $p=6.81\times10^{-7}$), EuroSAT ($-0.454$, $p=5.49\times10^{-10}$), and Oxford Pets ($-0.352$, $p=2.70\times10^{-6}$). Flowers-102 shows no significant conditional association ($\rho=+0.087$, $p=0.263$). Conversely, LogME remains strongly negatively associated with probe variability after controlling for Shesha\textsubscript{FS} on all six datasets, with partial correlations ranging from $-0.507$ to $-0.939$.

These results indicate that task alignment is the more consistent correlate of probe subset-sensitivity, while feature-split geometric stability contributes complementary statistical information on most datasets. We therefore do not interpret Shesha\textsubscript{FS} as a universal predictor of downstream probe robustness.

\begin{table}[t!]
\caption{Probe subset-sensitivity results for $K=20$ and $K=100$ random feature halves. We report the raw Spearman correlation between Shesha\textsubscript{FS} and probe-accuracy standard deviation $\sigma_{\mathrm{probe}}$, the corresponding partial correlation controlling for LogME, and the reciprocal partial correlation between LogME and $\sigma_{\mathrm{probe}}$ controlling for Shesha\textsubscript{FS}. The $K=100$ analysis is used in the main text; $K=20$ is reported as a robustness check.}
\label{tab:probe_k20_k100}
\centering
\small
\begin{tabular}{llccc}
\hline
Dataset & $K$
& $\rho(\mathcal{S},\sigma)$
& $\rho(\mathcal{S},\sigma\mid\mathrm{LogME})$
& $\rho(\mathrm{LogME},\sigma\mid \mathcal{S})$ \\
\hline
\multirow{2}{*}{CIFAR-10}
 & 20  & -0.302 & -0.286 & -0.898 \\
 & 100 & -0.294 & -0.318 & -0.939 \\
\hline
\multirow{2}{*}{CIFAR-100}
 & 20  & -0.278 & -0.342 & -0.744 \\
 & 100 & -0.337 & -0.513 & -0.860 \\
\hline
\multirow{2}{*}{DTD}
 & 20  & -0.325 & -0.279 & -0.406 \\
 & 100 & -0.405 & -0.371 & -0.507 \\
\hline
\multirow{2}{*}{EuroSAT}
 & 20  & -0.589 & -0.274 & -0.639 \\
 & 100 & -0.690 & -0.454 & -0.723 \\
\hline
\multirow{2}{*}{Flowers-102}
 & 20  & +0.418 & +0.181 & -0.786 \\
 & 100 & +0.379 & +0.087 & -0.833 \\
\hline
\multirow{2}{*}{Oxford Pets}
 & 20  & -0.086 & -0.275 & -0.645 \\
 & 100 & -0.117 & -0.352 & -0.713 \\
\hline
\end{tabular}
\end{table}

The qualitative pattern is stable across the two values of $K$. The sign of the raw Shesha--variability association is unchanged on all six datasets: CIFAR-10, CIFAR-100, DTD, EuroSAT, and Oxford Pets are negative at both $K=20$ and $K=100$, whereas Flowers-102 is positive at both values. Conditioning on LogME likewise yields negative associations on the same five datasets at both $K$ values, with Flowers-102 remaining positive but becoming small and nonsignificant at $K=100$. The expanded number of sampled feature halves therefore reduces Monte Carlo uncertainty without changing the cross-dataset qualitative pattern.

\subsubsection{Secondary Controls for Mean Probe Accuracy}

We also examined two controls for differences in average probe performance. First, we normalized probe variability by mean probe accuracy using the coefficient of variation. At $K=100$, the association with Shesha\textsubscript{FS} remains negative on CIFAR-10 ($\rho=-0.274$), CIFAR-100 ($-0.167$), DTD ($-0.285$), EuroSAT ($-0.681$), and Oxford Pets ($-0.110$), while remaining positive on Flowers-102 ($+0.384$). Second, we computed partial Spearman correlations controlling directly for mean probe accuracy. These correlations are negative on all six datasets at $K=100$, although the associations on Flowers-102 ($\rho_{\mathrm{partial}}=-0.073$, $p=0.346$) and Oxford Pets ($-0.138$, $p=0.074$) are not significant. The corresponding associations are stronger on CIFAR-10 ($-0.458$), CIFAR-100 ($-0.570$), DTD ($-0.533$), and EuroSAT ($-0.574$).

These analyses show that the cross-dataset pattern is not reducible to mean probe accuracy alone. We nevertheless treat them as secondary controls because mean accuracy does not capture task alignment as directly as LogME.

\subsubsection{Accuracy-Stratified Analysis}

As an additional descriptive check, we divided the 170 models within each dataset into terciles according to mean probe accuracy and recomputed the raw Shesha\textsubscript{FS}--$\sigma_{\mathrm{probe}}$ correlation within each tercile. At $K=100$, the association is negative in all three terciles on CIFAR-100, DTD, and EuroSAT. CIFAR-10 shows a stronger negative association in the middle tercile ($\rho=-0.453$, $p=4.49\times10^{-4}$) and weaker associations in the low- and high-accuracy terciles. Oxford Pets likewise shows a negative association only in the high-accuracy tercile ($\rho=-0.294$, $p=0.0265$). Flowers-102 is heterogeneous: the low-accuracy tercile is negative ($\rho=-0.343$, $p=0.009$), whereas the middle and high-accuracy terciles are positive ($\rho=+0.326$ and $+0.346$, respectively). 

The $K=20$ stratified results show the same broad dataset dependence. DTD is negative in all three terciles, EuroSAT is negative in all three with the middle tercile weaker, CIFAR-10 is strongest in the middle tercile, and Flowers-102 again changes sign between the low- and higher-accuracy groups. Because each tercile contains only 56--57 models and the within-tercile patterns vary substantially across datasets, we treat this analysis as descriptive rather than as evidence for a universal accuracy-dependent regime.

%%%%%%%%%%%%%%%%%%%%%%%%%%%%%%%%%%%%%%%%%%%%%%%%%%%%%%%%%%%%%%%%%%%%%%%%%%%
\section{Code Availability}
%%%%%%%%%%%%%%%%%%%%%%%%%%%%%%%%%%%%%%%%%%%%%%%%%%%%%%%%%%%%%%%%%%%%%%%%%%%

All custom code is available on GitHub (\url{https://github.com/prashantcraju/geometric-stability}, ~\citealt{github}). We have also released an open source Python library through PyPI (\url{https://pypi.org/project/shesha-geometry}; \citealt{shesha2026}).

% \newpage
\vskip 0.2in
\bibliography{sample}

\end{document}